\documentclass{article} 
\usepackage[final]{colm2025_conference} 

\usepackage{microtype}
\usepackage{hyperref}
\usepackage{url}
\usepackage{booktabs}
\usepackage{lineno}
\usepackage{paralist}

\newcommand{\quot}[1]{``#1''}
\usepackage{xspace}
\usepackage{soul}
\sethlcolor{pink}
\usepackage{multirow}
\newcommand{\LUMI}{LUMI\xspace}

\newcommand{\HI}{\textit{How-to-Instructions}\xspace}
\newcommand{\ID}{\textit{Interactive Discussion}\xspace}
\newcommand{\IN}{\textit{Informational Description}\xspace}

\newcommand{\LY}{\textit{Lyrical}\xspace}

\newcommand{\NA}{\textit{Narrative}\xspace}
\newcommand{\OP}{\textit{Opinion}\xspace}
\newcommand{\SP}{\textit{Spoken}\xspace}
\newcommand{\HIIN}{\textit{Instructive-Informational}\xspace}
\newcommand{\NE}{\textit{News}\xspace}
\newcommand{\DTP}{\textit{Description}\xspace}
\newcommand{\HIa}{\textit{How-to-Instructions} (HI)\xspace}
\newcommand{\IDa}{\textit{Interactive Discussion} (ID)\xspace}
\newcommand{\INa}{\textit{Informational Description} (IN)\xspace}
\newcommand{\IPa}{\textit{Informational Persuasion} (IP)\xspace}
\newcommand{\LYa}{\textit{Lyrical} (LY)\xspace}
\newcommand{\MTa}{\textit{Machine Translation} (MT)\xspace}
\newcommand{\NAa}{\textit{Narrative} (NA)\xspace}
\newcommand{\OPa}{\textit{Opinion} (OP)\xspace}
\newcommand{\SPa}{\textit{Spoken} (SP)\xspace}
\newcommand{\HIINa}{\textit{Instructive-Informational} (HI-IN)\xspace}
\newcommand{\NEa}{\textit{News} (ne)\xspace}
\newcommand{\DTPa}{\textit{Description} (dtp)\xspace}

\newcommand{\Mthree}{\textit{HI-IN-HI-dtp}\xspace}
\newcommand{\Mfour}{\textit{HI-IN-HI-dtp-OP}\xspace}
\newcommand{\Mfive}{\textit{HI-IN-HI-dtp-OP-NA}\xspace}
\newcommand{\Msix}{\textit{HI-IN-HI-dtp-OP-NA-ID}\xspace}
\newcommand{\Mseven}{\textit{HI-IN-HI-dtp-OP-NA-ID-SP}\xspace}

\usepackage[normalem]{ulem}

\usepackage{graphicx}

\definecolor{darkblue}{rgb}{0, 0, 0.5}
\hypersetup{colorlinks=true, citecolor=darkblue, linkcolor=darkblue, urlcolor=darkblue}

\title{Register Always Matters: Analysis of LLM Pretraining Data Through the Lens of Language Variation}

\author{Amanda Myntti, Erik Henriksson, Veronika Laippala \& Sampo Pyysalo  \\
University of Turku \\
\texttt{\{amanda.a.myntti,erik.henriksson,mavela,spyysalo\}@utu.fi} \\}

\begin{document}

\ifcolmsubmission
\linenumbers
\fi

\maketitle

\begin{abstract}

Pretraining data curation is a cornerstone in Large Language Model (LLM) development, leading to growing research on quality filtering of large web corpora. From statistical quality flags to LLM-based labelling systems, datasets are divided into categories, frequently reducing to a binary: those passing the filters are deemed as valuable examples, others are discarded as useless or detrimental. However, a more detailed understanding of the contribution of different kinds of texts to model performance is still largely lacking. In this article, we present the first study utilising \textit{registers} or \textit{genres}---a widely used standard in corpus linguistics to model linguistic variation---to curate pretraining datasets and investigate the effect of register on the performance of LLMs. 
We train small generative models with register classified data and evaluate them using standard benchmarks, and show that the register of pretraining data substantially affects model performance. We uncover surprising relationships between the pretraining material and the resulting models: using the \NE register results in subpar performance, and on the contrary, including the \OP class, covering texts such as reviews and opinion blogs, is highly beneficial.
While a model trained on the entire unfiltered dataset outperforms those trained on datasets limited to a single register, combining well-performing registers such as \HI, \IN, and \OP leads to major improvements.
Furthermore, analysis of individual benchmark results reveals key differences in the strengths and drawbacks of specific register classes as pretraining data: 
\HI excels at physical reasoning and sentence completion while barely crossing random baselines on world-knowledge benchmarks, while \NA boosts performance on social interaction tasks but struggles with scientific questions. These findings show that register is an important explainer of model variation and can facilitate more deliberate and detailed future data selection practices.
\end{abstract}

\section{Introduction}
\label{sec:Introduction}

Recent work on scaling laws and the limits of human-generated training data available for Large Language Model (LLM) training have caused an uptick in research on pretraining data quality as opposed to quantity \citep{kaplan2020scalinglawsneurallanguage,Hoffman-2022-chinchilla,Muenninghoff-2023-scaling,Villalobos-etal-2024-run-out-of-data}. While state-of-the-art models are still trained predominantly on large web corpora, there is increasing focus on filtering such data to remove potentially detrimental material. The selection of pretraining data often relies on simple heuristics or quality signals \citep{longpre-etal-2024-pretrainers,ostendorff2024llmdatasets}, sometimes together with LLM-created quality flags \citep{weber2024redpajama, henriksson2025finerweb10btrefiningwebdata}. However, language use varies along a number of dimensions other than the presumed dichotomy of positive or negative value for LLM training, ranging from persuasive, narrative, to informative texts. The effects of this variation on LLM capabilities remain unknown. In this paper, we approach pretraining data curation from a novel perspective: how do different registers (or genres) of training data affect LLM performance on commonly used benchmarks. 


\textit{Register studies} is a widely applied paradigm in corpus linguistics that examines language variation across different situations, from casual spoken conversations to persuasive and informational settings \citep{Biber_1988,Biber1995,biber2019register}. Registers are defined as situationally characterised text varieties, with typical classes encompassing news, reviews and song lyrics. Register has been shown to be one of the most important predictors of linguistic variation \citep{Biber-2012-linguistic-variation}---and, as put by the founder of register studies, Douglas Biber, \quot{\textit{register always matters}} \citep{Biber2013}.

Web register classification has long aimed to characterise different kinds of web texts in order to better utilise and understand the web as a corpus \citep{kilgarriff-grefenstette-2003-web-as-corpus,baroni-kilgarriff-2006-large}. However, early studies focused on pre-selected sets of registers presumed to be typical of the web, and thus lacked generalizability \citep{santini-2007-automatic-web-genre,sharoff-etal-2010-web-genre} and subsequently failed to capture the full range of linguistic variation found on the unrestricted web. 
\citet{egbert2015} addressed this gap by creating a corpus representing the full spectrum of English web registers, with \cite{laippala_ronnqvist2023} presenting the first deep learning-based web register classifier trained on the data and targeting the entire unrestricted web. Other studies expanded this approach to multilingual settings \citep{repo2021zeroshot,ronnqvist-etal-2021-multilingual,Kuzman-genre-identification-2023,henriksson2024automaticregisteridentificationopen}, showing that web registers are identifiable in web-scale data. These advances now enable the integration of corpus linguistics frameworks with LLM pretraining data curation---our study being to the best of our knowledge the first to bridge these areas, despite the well-documented role of registers in linguistic variation.

To understand the characteristics of web registers in the context of LLM pretraining, we train small generative models up to 100 billion tokens with web data 
with datasets curated to only contain texts from one register class.
We evaluate these models using well-known benchmarks and analyse the results through average accuracy and performance on individual benchmarks, revealing how each register impacts model capabilities. Our results show that register has a substantial impact on model performance. 
Additionally, we find that training on a combination of register-specific datasets that displayed strengths in different benchmarks can lead to major improvements:
registers like \OP boosts model performance, while incorporating examples from register \ID or \SP at the expense of \HI, \IN and \OP leads to performance drops. With these findings, our study aims to open up discussion about linguistic selection of training data, and to show that register classification holds potential as a tool for analysing pretraining data.

\section{Related work}
\label{sec:Related-work}

\label{sec:training-data-curation}

Ablation studies, where small generative models are trained with different parts of a large dataset and evaluated on downstream tasks, are a well-established method for evaluating dataset quality and optimising data curation methods.

\cite{gao2020pile} validated the quality of their Pile datasets by evaluating models trained on the Pile, CommonCrawl, and CC-100. \cite{longpre-etal-2024-pretrainers} pretrain small models on data filtered by different qualities (data age, data domain, data toxicity) and analyse the effects of this curation via model evaluation. Similarly, \citet{sachdeva2024traindataefficientllms} use an ablation framework to find that diversity and coverage of different topics in the pretraining data are crucial to models that succeed in multiple benchmarks. Likewise, \citet{Burchell-HPLT2} released a multilingual dataset and performed ablation studies to measure the quality of different versions of their datasets. \cite{henriksson2025finerweb10btrefiningwebdata} use an LLM to line-by-line classify training documents to optimise dataset composition and evaluate with a downstream task. 

The methodology of our approach most closely resembles that of the FineWeb ablation studies \citep {penedo2024finewebdatasetsdecantingweb}. 
The FineWeb dataset was created by conducting analyses of tens of heuristic filters to select the data curation methods that result in the best pretraining data with respect to the downstream performance of the trained model. Currently, datasets produced by the FineWeb pipeline are available for over a thousand languages \citep{penedo2024fineweb-2}. 

Current leading work on dataset composition optimisation includes \citet{su2024nemotroncctransformingcommoncrawl} and \citet{li2024datacomplmsearchgenerationtraining}.
\citet{lee-etal-2022-deduplicating} established deduplication as a standard procedure in pretraining data selection pipelines. Other common approaches for selecting pretraining data from web sources include URL-based filtering (e.g. \citet{penedo-2023-refinedweb}), text statistics such as word count, repetition, or presence of blacklisted words (e.g. \citet{raffel-2020-transferlearning}), filtering based on perplexity (e.g. \citet{Muenninghoff-2023-scaling},\citet{weber2024redpajama}), and selection based on similarity against a selected good-quality corpus (e.g. \citet{brown-etal-2020-fewshot}, \citet{tirumala-etal-2023-improving-pretraining}).
According to a survey by \citet{albalak-2024-survey}, it remains an open question whether quality filtering data improves model performance, and in which cases: \citet{dodge-etal-2021-documenting} show that url-based pretraining data filtering leads to bias in the resulting data, while reference corpora used for similarity selection can be biased in language, topics, or demographics \citep{rae2022scalinglanguagemodelsmethods, gururangan-etal-2022-whose}.

While register studies is a widely applied standard to analyse language variation in corpus linguistics, the term \textit{genre} is more often used in the NLP field \citep{sharoff-etal-2010-web-genre,petrenz-webber-2011-squibs,sharoff-2020-know,kuzman-etal-2023-get}. However, both approaches target text categories such as opinionated, informational, and lyrical.

Register studies has a long tradition in mapping the relationship between typical linguistic features associated with texts and different situational contexts where language is used, in both online and offline settings (e.g., \citet{Biber1995,Biber_Egbert_2018,staples2025}). \cite{Biber-2012-linguistic-variation} even shows that register is one of the most important predictors of linguistic variation. Similarly, web register (or genre) classification is a widely applied approach to classify web content \citep{kilgarriff-grefenstette-2003-web-as-corpus, santini-2007-automatic-web-genre,petrenz-webber-2011-squibs,myntti-etal-2024-intersecting}. 

Since the release of the first web register corpus covering the full unrestricted web \citep{egbert2015}, automatic register classification using machine learning has been successfully applied to the unrestricted web, achieving nearly human-level performance \citep{laippala_ronnqvist2023, kuzman-etal-2023-get}. \citet{henriksson2024automaticregisteridentificationopen} also show that register characteristics transcend language boundaries, allowing classification models to identify registers even in languages that were not present in the training data (zero-shot languages). These advances enable the current study to select pretraining data by register. Our approach also aligns with calls for greater transparency in foundation models \citep{bommasani2023foundationmodeltransparencyindex}, as it provides structured metadata about the linguistic composition of training data, helping researchers and users to understand what types of texts a model has been exposed to during training.

\begin{table}[!t]
\small
\begin{tabular}{@{}llc@{}}
\toprule
Individual registers                  & Description                                          & Available     \\ \midrule
\hspace{5pt} \HIa                         & Recipes, instructions                       & 100 B         \\
\hspace{5pt} \IDa                         & Discussion forums        & 314 B         \\
\hspace{5pt} \INa                         & Wikis, information sites                  & 695 B         \\
\hspace{5pt} \IPa                      & Advertisements, commercial sites            & 421 B         \\
\hspace{5pt} \LYa                         & Song lyrics, poems                       & 20 B          \\
\hspace{5pt} \MTa                        & Machine translation                                  & 306 B         \\
\hspace{5pt} \NAa                         & News, blogs                              & 545 B         \\
\hspace{5pt} \OPa                         & Opinionated texts, religious sermons       & 416 B         \\
\hspace{5pt} \SPa                         & Interviews, speeches                        & 32 B          \\ 
\hspace{5pt} \HIINa                      & Hybrid; Documents predicted as both HI and IN                & 70B           \\
\hspace{5pt} \NEa                         & Subregister of NA; news                              & 404 B         \\
\hspace{5pt} \DTPa                        & Subregister of IN; decription of a thing or a person & 781 B         \\ \midrule
Combinations              &                                                      &               \\ \midrule
\hspace{5pt} \Mthree      & 1/3rd sampled from HI-IN, HI, dtp                                 & -- \\ 
\hspace{5pt} \Mfour    & 1/4th sampled from HI-IN, HI, dtp, OP                              & -- \\
\hspace{5pt} \Mfive  & 1/5th sampled from HI-IN, HI, dtp, OP, NA                                 & -- \\ 
\hspace{5pt} \Msix   & 1/6th sampled from HI-IN, HI, dtp, OP, NA, ID                                 & -- \\ 
\hspace{5pt} \Mseven  & 1/7th sampled from HI-IN, HI, dtp, OP, NA, ID, SP                                 & -- \\ \midrule 
Baselines & & \\ \midrule
\hspace{5pt} HPLT v2 deduplicated & \cite{Burchell-HPLT2} & -- \\
\hspace{5pt} FineWeb & \cite{penedo2024finewebdatasetsdecantingweb} & -- \\ \bottomrule
\end{tabular}
\caption{Datasets used in our experiments. Each dataset corresponds to one trained model. The \textit{Individual registers} section describes the main register labels and hybrid \& subregisters. The \textit{Combinations} section contains combined register classes, and \textit{Baselines} lists the baseline datasets we use in our experiments. The \textit{Available} column contains the approximate token counts available in the HPLT v2 deduplicated corpus.}
\label{tab:register-descriptions}
\end{table}

\section{Methods}
\label{sec:methods}

\subsection{Data}
\label{sec:Data}

We use HPLT version 2.0\footnote{Available at \url{https://hplt-project.org/datasets/v2.0}} \citep{Burchell-HPLT2} as our source data. This collection includes datasets for 193 languages and additional parallel datasets for 50 languages paired with English. HPLT v2 datasets have been processed from a combination of Internet Archive and Common Crawl using Trafilatura \citep{barbaresi-2021-trafilatura} with language identification performed using OpenLID \citep{burchell-etal-2023-open}. Two versions of each dataset are provided, \textit{deduplicated}, with MinHash deduplication \citep{Broder-1998-minhash}, and \textit{cleaned}, with additional cleaning heuristics applied.
An evaluation by \citet{Burchell-HPLT2} showed the cleaned English dataset to be of high quality, with models trained on the data achieving performance comparable to ones trained on the FineWeb dataset \citep{penedo2024finewebdatasetsdecantingweb}. In this study, we nevertheless chose to use the deduplicated English dataset, which yields slightly worse performance in LLM pretraining. This choice allows us to examine the effect of register with minimal interference from cleaning procedures that might disproportionately affect certain registers and thus bias our analysis. This decision is also motivated by the uneven distribution of register labels in web data; with fewer cleaning steps, we retain more data, which is especially important for the less frequent register classes.

The HPLT v2 dataset includes predictions of register classes as probabilities for each document, generated using the multi-label register classifier by \citet{henriksson2024automaticregisteridentificationopen}. This classifier was fine-tuned from XLM-RoBERTa-Large \citep{conneau-etal-xlm-roberta} using the multilingual CORE corpus introduced in the same study. The CORE register scheme is hierarchical, with 9 main registers divided into 25 subregisters that further describe the content of each document. Documents can have none, one, or multiple main registers (referred to as \textit{hybrids}), along with any associated subregister labels. The full register scheme can be found in \citet{henriksson2024automaticregisteridentificationopen}, and example documents are given in Appendix~\ref{sec:appendix-text-examples}.

Although the register labels are available for multiple languages, we limit our analysis to English due to the availability of varied and well-established benchmarks.
For our experiments, we assign register labels to documents from the given probabilities using a classification threshold of 0.4, optimised for English. From the register classes, we select all main registers, 2 subregisters, and one hybrid for our experiments. We included the subregisters and hybrids for specific reasons: the subregisters \NE and \DTP are among the largest subregister classes, each with over 100 billion tokens available---sufficient data to train models and draw meaningful comparisons with their parent registers \NA and \IN. Moreover, there has been particular interest in using news articles (e.g. \citet{delarosa-2025-impactcopyrightedmateriallarge}) and web encyclopedias (i.a. Wikipedia) to train LLMs, which correspond to classes \NE and \DTP in the register scheme. The hybrid class \HIIN was included to study the differences between individual registers and hybrids, based on preliminary results showing that register-specific models trained on these classes perform well overall but differ across individual benchmark tasks. The selected register classes and their abbreviations are described with examples in Table~\ref{tab:register-descriptions}. Subregister classes not selected for our experiments are present in the data; however, they merged to their main-level class.

Using these labels, we sample 100 billion tokens per register, using all available data for registers with fewer than 100 billion tokens.  We limit our analysis to documents over 200 characters and remove documents with extremely high token counts (corresponding to over 300,000 words\footnote{Defined as white-space and punctuation separated segments}) due to hardware constraints. These excluded documents represent less than 0.3\% of the total data and primarily belong to the \IN class.

We also experiment with training on combinations of certain registers, which are listed in Table~\ref{tab:register-descriptions} under \quot{Combinations}. These combination datasets are created by sampling equally from individual register datasets: for example, in the combination dataset \Mthree, 1/3rd of the training samples are from \HIINa,  1/3rd from \HIa, and 1/3rd from \DTPa, with our dataset creation step ensuring no duplicates are present in the combined data. See Section~\ref{sec:experiments} for further information and motivation for these datasets.

\subsection{Model training}
\label{sec:model_training}

We replicate the training setup of \citet{penedo2024finewebdatasetsdecantingweb}, using the same model architecture (Llama), model size (1.71 billion parameters), and other training parameters. In total, the models are trained up to 100 billion tokens. See Appendix \ref{sec:appendix-training-setup} for further information on the training setup. As a baseline, we use a model trained on a random sample of the HPLT~v2 deduplicated data. This allows us to compare the effects of using a single register as opposed to using data with the natural distribution of registers in the source dataset. 
Finally, as our setup mirrors that of \citet{penedo2024finewebdatasetsdecantingweb}, we also include results for a model trained on a sample of the FineWeb dataset as a further point of comparison and as a reference for interpreting our findings.

\subsection{Evaluation}
\label{sec:evaluation}

To maintain comparability with previous work, we also follow
\cite{penedo2024finewebdatasetsdecantingweb} in our evaluation. Specifically, we use the \texttt{LightEval} \citep{Fourrier-lighteval} evaluation harness with the following benchmarks in a zero-shot setting for all tasks: 

\begin{compactitem}

\item\textbf{HellaSwag} \citep{zellers2019hellaswag} contains logical sentence completion tasks in the form of presenting a paragraph and an incomplete sentence with alternative sentence endings. 
\item\textbf{WinoGrande} \citep{sakaguchi-2021-winogrande} evaluates commonsense reasoning in a binary classification setting using pronoun resolution problems. 
\item\textbf{PIQA} \citep{Bisk2020piqa} focuses on physical commonsense reasoning, with examples involving manipulation of physical objects. 
\item\textbf{SIQA} \citep{sap-etal-2019-social} contains tasks on reasoning about social situations. 
\item\textbf{OpenBookQA} \citep{OpenBookQA2018} focuses on tasks requiring multi-step reasoning with additional context given. 
\item\textbf{ARC Easy and ARC Challenge} \citep{clark-2018-ARC} are multiple choice benchmarks with science exam questions. 
\item\textbf{CommonsenseQA} \citep{talmor-etal-2019-commonsenseqa} consists of multiple-choice questions requiring prior world knowledge. 
\item\textbf{MMLU} \citep{hendrycks2021ethics,hendryckstest2021} is a multiple-choice question answering dataset covering 57 distinct categories, such as US politics and anatomy.
\end{compactitem}

These benchmarks were chosen by \citet{penedo2024finewebdatasetsdecantingweb} due to their compatibility with small model size, showing reliable, low-variance and monotonic signals even with limited training data, which is crucial for our experiments.  
For MMLU, we consider all 57 categories as one task and average our results accordingly. This ensures that models reaching high accuracy on MMLU but mediocre results on other benchmarks do not dominate the average results. See Appendix~\ref{sec:appendix_numerical} for further discussion. 

The selected benchmarks reflect the most common present-day expectations placed on LLMs: fluency, scientific and real-world knowledge, problem solving, and reasoning. 
However, this selection inevitably disadvantages certain register models. For example, models trained on the \LY register are likely to show poor performance, as none of the benchmarks measure poetic capabilities that would be expected from a model trained on lyrical data.  This does not mean that lyrical capabilities are unwanted or unnecessary for LLMs, but rather highlights the importance of benchmark diversity and the significance of selecting benchmarks that measure abilities relevant to specific use cases. In our case, although the benchmark selection disadvantages some registers and their expected performance more than others, using the full range of registers covering all kinds of web texts yields a complete picture of registers' effects on LLM pretraining.

\section{Experiments}
\label{sec:experiments}

Our experiments are divided into two parts: First, we examine the effects of register on model performance by training models for the 9 main-level registers, one hybrid class (\HIIN), and two subregister classes (\NE and \DTP), presented in Table \ref{tab:register-descriptions} under \quot{Individual registers}. These experiments are meant to highlight the differences of registers as pretraining data.  

Second, we investigate the effect of further mixing registers by selecting top-performing registers from our first experiment. This allows us to observe how adding or removing specific registers affects overall model performance and whether the strengths of individual registers are preserved in combined models. The selected top-scoring register combinations are presented in Table \ref{tab:register-descriptions} under \quot{Combinations}. When sampling data for register combinations, we ensure that no duplicate documents are included, even when combining the hybrid class \HIINa with the register \HIa. We chose to use the subregister \DTPa instead of the main-level register \INa because \IN showed signs of performance degradation in later training steps, and as noted in Section \ref{sec:Data}, we had to exclude some documents from the \IN dataset. To maintain comparability, we train all models up to 100 billion tokens, although the larger amount of combined data would allow for longer training. Our aim is not to identify optimal register combinations, but rather to explore how register-based data curation might influence model performance on different tasks.

Some of our datasets contain fewer than 100 billion tokens, which necessitates repeating data during training. This raises questions about the validity of our results, as \citet{Hoffman-2022-chinchilla} argue that repeating data can lead to performance degradation. However, \citet{taylor2022galacticalargelanguagemodel} report validation loss continues to improve up to 4 epochs for a 120 billion parameter model, and \citet{Muenninghoff-2023-scaling} find that for the GPT-2 \citep{radford2019gpt2} architecture with 2-8 billion parameters, multi-epoch training has negligible effects when limited to 4 or fewer epochs. \citet{Xue-2023-insights-from-scaling} further demonstrate that smaller models are less susceptible to the negative effects of repeated training examples in T5-architecture models \citep{raffel-2020-transferlearning}. Therefore, we are reasonably confident in the results from models trained on \HIIN (1.4 epochs) and \SP (3.1 epochs). The model trained on our smallest dataset, \LY (5 epochs), performs poorly even within the first epochs, suggesting its low performance is due to register characteristics and benchmark selection rather than data repetition. None of the combination models exceed 1 epoch of training, since we sample the 100 billion tokens from combinations of multiple datasets, together exceeding 100 billion tokens.

We additionally compare our register scheme to three other quality classifiers, FineWeb-edu classifier \citep{penedo2024finewebdatasetsdecantingweb}, the DCLM quality classifier \citep{li2024datacomplmsearchgenerationtraining}, and NVIDIA's NemoCurator Quality Classifier DeBERTa \citep{he2023debertav3improvingdebertausing}. These experiments and their results are presented in Appendix~\ref{sec:appendix-comparison}.

\section{Results}
\label{sec:results}


\subsection{Average performance}
\label{sec:results-average}

\begin{figure}
    \small
    \centering
    \includegraphics[width=0.9\linewidth]{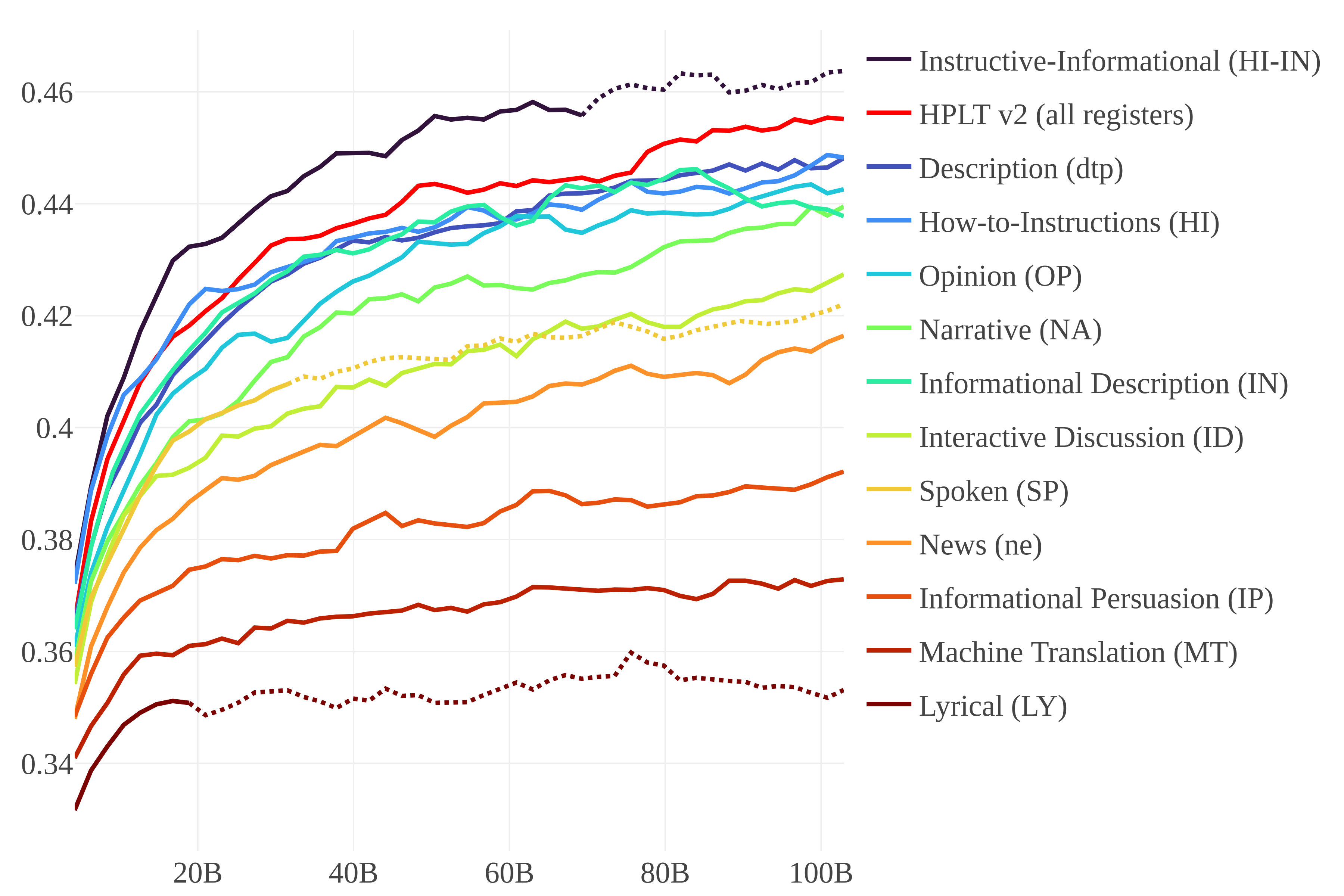}
    \caption{Individual register and HPLT v2 (deduplicated) model accuracies as a function of the number of training tokens. Rolling average over 6 billion tokens (3 adjacent checkpoints) applied for ease of reading overlapping lines, see Table~\ref{tab:final_numerical} for numerical results for the final checkpoint. Dotted lines indicate training continuing over one epoch, legend in order of last checkpoint performance.}
    \label{fig:results-individual}
\end{figure}

\begin{figure}
    \small
    \centering
    \includegraphics[width=0.9\linewidth]{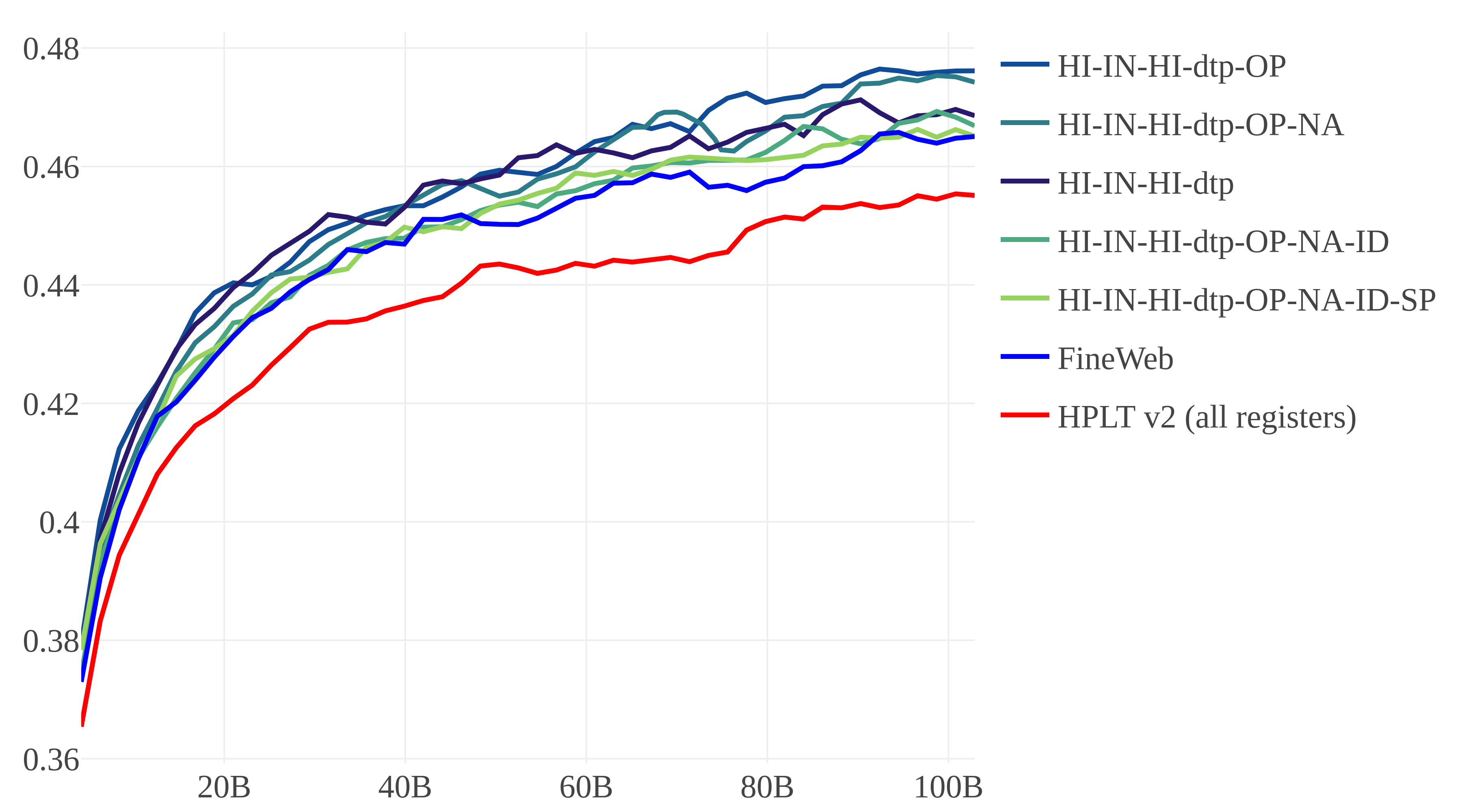}
    \caption{Combination model and baseline accuracies as a function of the number of training tokens. Rolling average over 6 billion tokens (3 adjacent checkpoints) applied for ease of reading overlapping lines, see Table~\ref{tab:final_numerical} for numerical results for the final checkpoint. Legend in order of last checkpoint performance.}
    \label{fig:results-combinations}
\end{figure}

The results for individual register models, averaged over all benchmarks, are presented in Figure~\ref{fig:results-individual} and Table~\ref{tab:final_numerical}, and detailed in Appendix~\ref{sec:appendix_numerical}.
The evaluation shows striking differences in accuracy between the models, demonstrating that the register of the pretraining data has a substantial effect on model performance.
We find that the model trained on the full HPLT~v2 data outperforms models trained on individual registers, demonstrating that a mix of registers is important for LLM training.
The performance ranking of individual registers generally aligns with intuitions about their content and what is typically considered high-quality pretraining data: registers like \LYa, \MTa, and \IPa yield worse-performing models, while models trained on registers containing informational content and instructions, such as \DTPa and \HIa, perform the best. 
Surprisingly, \OPa yields the 4th best performing model. Both \INa and \NAa models perform worse than \HI, \DTP and \OP. Following these, with a noticeable performance drop, are the models trained on \IDa, \SPa and remarkably, the subregister \NEa, consisting of news articles, which performs among the worst.

As opposed to the main register models, 
the hybrid class model \HIINa
outperforms the model trained on the full HPLT~v2 data.
This shows that hybrid documents do not simply result in averaged performance between the two hybridised registers, but can show better performance than either alone. This finding motivated our second experiment, investigating register combinations.
The subregisters \NEa (subregister of \NA) and \DTPa (subregister of \IN) show divergent patterns when compared to their parent registers: \NE drastically underperforms its main-level register \NA, while \DTP achieves higher accuracy than \IN. 
We analyse differences between main-subregister pairs in more detail in the following section.

\begin{table}[]
\small
\centering
\begin{tabular}{lcllc}
\toprule
Pretraining dataset    & Accuracy &  & Pretraining dataset & Accuracy \\ \midrule 
\Mfour  & 0.475    &  & \OPa & 0.447    \\
\Mfive  & 0.472    &  & \NAa & 0.441    \\
\Mthree & 0.466    &  & \INa & 0.437    \\
FineWeb                & 0.466    &  & \IDa & 0.431    \\
\Mseven & 0.465    &  & \SPa & 0.422    \\
\HIINa  & 0.465    &  & \NEa & 0.418    \\
\Msix   & 0.464    &  & \IPa & 0.393    \\
HPLT v2 (deduplicated, all registers) & 0.457    &  & \MTa & 0.374    \\
\DTPa   & 0.452    &  & \LYa & 0.358    \\
\HIa    & 0.447    &  &                     &          \\ \midrule
\end{tabular}
\caption{Results by the last checkpoint performance in numerical format. Note that the values differ slightly from the Figures~\ref{fig:results-individual} and \ref{fig:results-combinations} due averaging over adjacent checkpoints in the figures.}
\label{tab:final_numerical}
\end{table}

\begin{figure}[!t]
    \small
    \includegraphics[width=0.87\linewidth]{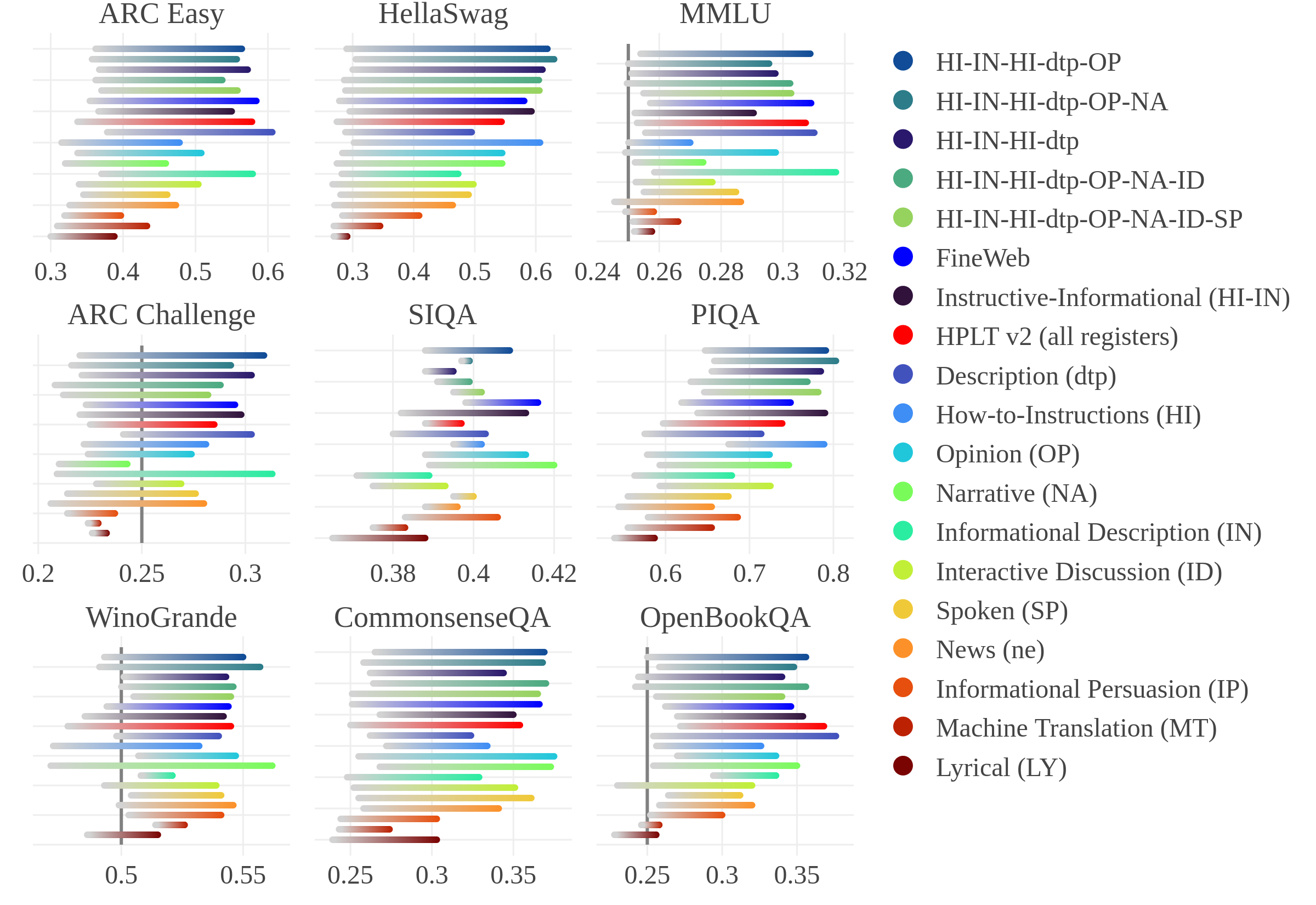}
    \caption{Change of accuracy from first to final checkpoint on individual benchmarks shown as a range, with grey indicating the first checkpoint and colours indicating the last checkpoint. The random-guess threshold is shown as a grey vertical line in cases where at least one model falls below it. Bars and legend shown in order of average accuracy.}
    \label{fig:results-per-task-range}
\end{figure}

As the results show, pretraining a model with data from only one register class does not provide improvements compared to training with the full dataset. This aligns with the established notion that pretraining data should have variability to yield capable models. The results for our combination model experiment are shown in Figure~\ref{fig:results-combinations}, Table~\ref{tab:final_numerical}, and again with more detail in Appendix \ref{sec:appendix_numerical}. The results clearly support the above idea: all combination models outperform the individual register models and the baseline HPLT v2 model.
Additionally, the best models in our study (\Mfour and \Mfive) reach higher average accuracy than the models trained on the HPLT v2 and FineWeb datasets. 
\Mthree model shows much worse performance than \Mfour, which further includes \OPa. This suggests that opinionated texts provide valuable pretraining data and are key to high model performance, at least as measured by the selected benchmarks. Further adding the \NAa dataset, as seen in the \Mfive model, slightly worsens performance, though the difference is small. 
Significant drops occur when including \IDa and \SPa in models \Msix and \Mseven, indicating either that these registers contain qualities that harm the abilities measured on these benchmarks, or that they do not provide benefits that outweigh the reduced proportion of \HIINa, \DTPa, and \OPa in the training mix.

\subsection{Per task}
\label{sec:per-task}

The results for each benchmark separately are shown in 
Figure \ref{fig:results-per-task-range} and detailed further in Appendix~\ref{sec:app-detailed}.
The differences between benchmarks are remarkable, both in the performance ranking of models and in the way accuracy trends with respect to the number of training tokens. This shows that registers have strengths in different tasks. 
Most benchmarks display the monotonicity condition that influenced the selection of \citet{penedo2024finewebdatasetsdecantingweb}, as they tend towards increased accuracy with more training tokens. A clear outlier is the SIQA benchmark, where increases in accuracy are very small, with the largest increase seen in \NAa and \HIINa models, both increasing approximately by 0.03 during the training. 

Generally, the trends observed in Section \ref{sec:results-average} hold. The models with the lowest average accuracy---\LYa, \MTa and \IPa---perform poorly on all benchmarks, and, notably, worse than a random guess on ARC Challenge.
The model for subregister \NEa, despite weak average performance, improves on the MMLU and WinoGrande benchmarks, while \SPa enhances results on CommonsenseQA. The differences between benchmarks become more pronounced with models trained on \NAa, \INa, \OPa, \HIa, and \IDa, which display great variation in ranking across tasks, highlighting their strengths. The \NA model performs well on SIQA and WinoGrande but barely surpasses the random baseline on ARC Challenge.  \IN trumps other models on MMLU and both ARC benchmarks, but ranks low to mediocre elsewhere.
The model trained on the \HI dataset outperforms almost all other models on PIQA and HellaSwag. 
This might be because PIQA measures reasoning in physical tasks, which aligns with the instructions related to physics included in \HI.  On the other hand, \HI ranks among the worst on MMLU and CommonsenseQA.
The \OP model excels in WinoGrande, SIQA and CommonsenseQA. This signal also appears in the combination models' performance on CommonsenseQA: including \OP in the training data greatly increases accuracy on this benchmark, unlike the \Mthree model, which excludes \OP.  

The hybrid \HIINa model generally performs between individual \HIa and \INa models, and is notably closer in performance to whichever yields better accuracy. Exceptions to this pattern are OpenBookQA, CommonsenseQA, and WinoGrande, where the \HIIN model shows better accuracy than either \HI or \IN separately. 
Specifically in the case of WinoGrande, this is surprising, as both \HI and \IN struggle with this benchmark. From this result, we conclude that hybrids may contain qualities not found in individual registers and require further study. The \NEa model trails behind \NAa, except on the MMLU and ARC Challenge benchmarks. 
This is likely due to news articles covering topics related to MMLU tasks, such as politics, business, science, and the environment. Similarly, the other main-subregister pair, \DTPa and \INa, shows differences in performance across benchmarks, with the \DTP-model outshining \IN in ARC Easy, HellaSwag, PIQA, WinoGrande, and OpenbookQA. \IN, on the other hand, excels in ARC Challenge and MMLU. Specifically for the WinoGrande benchmark, the performance difference might be explained by the \DTP register containing more texts focused on things and people, thus resulting in better pronoun-resolution abilities. 

The combination models lead in performance on most benchmarks. 
On MMLU specifically, we observe a huge drop between the \Mfour model and others: including \OP in the pretraining data increases performance, while removing \OP or further adding \NA greatly disrupts model performance on this benchmark. Notably, the models \Mfour and \Mfive are trained with registers that individually perform well on a wide range of benchmarks: \HIIN on HellaSwag and ARC Easy, \HI on Hellaswag and PIQA, \DTP on ARC Easy, ARC Challenge, and OpenBookQA, \OP on CommonsenseQA, and \NA on SIQA and WinoGrande.

\section{Conclusion}
\label{sec:conclusions}

In this paper, we presented the first study investigating the effect of linguistic register on the performance of small generative LLMs. Our findings show that register is an important explainer of LLM performance, and we were able to show surprising relationships between the register of pretraining data and model accuracy on different benchmarks.
While using the whole dataset yielded better performance than limiting the pretraining data to any individual register class,
we found specific combinations of registers that outperform the full uncurated dataset. These findings can be used to increase pretraining data transparency and to optimise data selection methods, potentially addressing specific performance gaps by incorporating appropriate registers.

\section*{Limitations and future work}

As a first study in the intersection of two broad lines of research -- the linguistic registers and LLM training -- our work leaves a number of questions incompletely answered. Our experiments are conducted in English only. While register classification has shown multilingual prowess, our results might not generalise to other languages, and we leave multilingual analysis for future work. 
The benchmarks used, though standard in studies of this type, do not fully measure all model capabilities, as discussed regarding the \textit{Lyrical} class. Evaluating with more diverse benchmarks would provide more comprehensive results and might reveal strengths and drawbacks we were unable to uncover.

Bias from data curation methods is known to propagate to trained models \citep{rae2022scalinglanguagemodelsmethods, gururangan-etal-2022-whose}, and register classification is not exempt from this issue. To mitigate bias, we used the \textit{deduplicated} version of the HPLT v2 data. Although this version has been processed with Trafilatura, \citet{penedo2024finewebdatasetsdecantingweb} note that Trafilatura can leave \quot{undesirable} content in the dataset, potentially affecting some registers more than others. The possibility of benchmark contamination also exists; register classification could systematically assign leaked benchmark material to specific classes, potentially explaining some performance differences.

\ifcolmsubmission
\else 
\section*{Acknowledgments}
This project has received funding from the \textit{Finnish Doctoral Program Network in Artificial Intelligence, AI-DOC} under decision number VN/3137/2024-OKM-6, the European Union’s \textit{Horizon Europe research and innovation programme} under grant agreement No. 101070350, the \textit{Digital Europe Programme} under grant agreement No. 101195233, and the Research Council of Finland under grant No. 362459.
The contents of this publication are the sole responsibility of its authors and do not necessarily reflect the opinion of the European Union. We wish to acknowledge CSC - IT Center for Science, Finland, for computational resources.
\fi

\section*{Ethics Statement}

Training large language models is resource-demanding. This fact guided our decisions: we used small model sizes and trained only up to the point where clear conclusions can be drawn. In total, we used approximately 11,000 GPU hours per model. See Appendix \ref{sec:appendix-training-setup} for calculation of the total computational cost. To further mitigate the environmental impact of training, we used \LUMI supercomputer, which is among the most environmentally friendly supercomputers in the world\footnote{\url{https://lumi-supercomputer.eu/sustainable-future/}}. The models trained for this study are planned to be used in future research.

Releasing our models publicly for reproducibility opens up the possibility of our models being used in contexts other than research. Although we do not evaluate the possible toxicity of the generations they produce, the model sizes are small, and they are trained only up to 100 billion tokens, which makes their generative capabilities weak and thus likely unsuitable for applications outside the scope of this study.

\section*{Reproducibility statement}

We have prioritised reproducibility in our work and make all artefacts created as part of this study, including code, model training configurations, and the trained model checkpoints, available on \url{https://huggingface.co/TurkuNLP/register-ablations} under open licenses. Our work only makes use of data openly available from \url{https://hplt-project.org/datasets/v2.0}, and our model training setup is detailed in Appendix~\ref{sec:appendix-training-setup}.

\bibliography{ref}

\begin{thebibliography}{65}
\providecommand{\natexlab}[1]{#1}
\providecommand{\url}[1]{\texttt{#1}}
\expandafter\ifx\csname urlstyle\endcsname\relax
  \providecommand{\doi}[1]{doi: #1}\else
  \providecommand{\doi}{doi: \begingroup \urlstyle{rm}\Url}\fi

\bibitem[Albalak et~al.(2024)Albalak, Elazar, Xie, Longpre, Lambert, Wang, Muennighoff, Hou, Pan, Jeong, Raffel, Chang, Hashimoto, and Wang]{albalak-2024-survey}
Alon Albalak, Yanai Elazar, Sang~Michael Xie, Shayne Longpre, Nathan Lambert, Xinyi Wang, Niklas Muennighoff, Bairu Hou, Liangming Pan, Haewon Jeong, Colin Raffel, Shiyu Chang, Tatsunori Hashimoto, and William~Yang Wang.
\newblock A survey on data selection for language models, 2024.
\newblock URL \url{https://arxiv.org/abs/2402.16827}.

\bibitem[Ansarifar et~al.(2025)Ansarifar, Shahriari, Staples, and Ghazanfari]{staples2025}
Ahmad Ansarifar, Hesamoddin Shahriari, Shelley Staples, and Mohammad Ghazanfari.
\newblock Linguistic variation in two written academic sub-registers.
\newblock \emph{Revista Española de Lingüística Aplicada/Spanish Journal of Applied Linguistics}, 38\penalty0 (1):\penalty0 162--191, 2025.
\newblock ISSN 0213-2028.
\newblock \doi{https://doi.org/10.1075/resla.22052.ans}.
\newblock URL \url{https://www.jbe-platform.com/content/journals/10.1075/resla.22052.ans}.

\bibitem[Barbaresi(2021)]{barbaresi-2021-trafilatura}
Adrien Barbaresi.
\newblock Trafilatura: {A} web scraping library and command-line tool for text discovery and extraction.
\newblock In Heng Ji, Jong~C. Park, and Rui Xia (eds.), \emph{Proceedings of the 59th Annual Meeting of the Association for Computational Linguistics and the 11th International Joint Conference on Natural Language Processing: System Demonstrations}, pp.\  122--131, Online, August 2021. Association for Computational Linguistics.
\newblock \doi{10.18653/v1/2021.acl-demo.15}.
\newblock URL \url{https://aclanthology.org/2021.acl-demo.15/}.

\bibitem[Baroni \& Kilgarriff(2006)Baroni and Kilgarriff]{baroni-kilgarriff-2006-large}
Marco Baroni and Adam Kilgarriff.
\newblock Large linguistically-processed web corpora for multiple languages.
\newblock In \emph{Demonstrations}, pp.\  87--90, 2006.
\newblock URL \url{https://aclanthology.org/E06-2001/}.

\bibitem[Biber(1988)]{Biber_1988}
Douglas Biber.
\newblock \emph{Variation across Speech and Writing}.
\newblock Cambridge University Press, 1988.

\bibitem[Biber(1995)]{Biber1995}
Douglas Biber.
\newblock \emph{Dimensions of Register Variation: A Cross-Linguistic Comparison}.
\newblock Cambridge University Press, Cambridge, 1995.

\bibitem[Biber(2012)]{Biber-2012-linguistic-variation}
Douglas Biber.
\newblock Register as a predictor of linguistic variation.
\newblock \emph{Corpus Linguistics and Linguistic Theory}, 8\penalty0 (1):\penalty0 9--37, 2012.
\newblock \doi{doi:10.1515/cllt-2012-0002}.
\newblock URL \url{https://doi.org/10.1515/cllt-2012-0002}.

\bibitem[Biber(2013)]{Biber2013}
Douglas Biber.
\newblock Interview with {D}ouglas {B}iber.
\newblock \emph{Journal of English Linguistics}, 41\penalty0 (4):\penalty0 359--379, 2013.
\newblock \doi{10.1177/0075424213502237}.
\newblock URL \url{https://doi.org/10.1177/0075424213502237}.

\bibitem[Biber \& Conrad(2019)Biber and Conrad]{biber2019register}
Douglas Biber and Susan Conrad.
\newblock \emph{Register, Genre, and Style}.
\newblock Cambridge University Press, Cambridge, 2019.

\bibitem[Biber \& Egbert(2018)Biber and Egbert]{Biber_Egbert_2018}
Douglas Biber and Jesse Egbert.
\newblock \emph{Register Variation Online}.
\newblock Cambridge University Press, 2018.

\bibitem[Bisk et~al.(2020)Bisk, Zellers, Bras, Gao, and Choi]{Bisk2020piqa}
Yonatan Bisk, Rowan Zellers, Ronan~Le Bras, Jianfeng Gao, and Yejin Choi.
\newblock Piqa: Reasoning about physical commonsense in natural language.
\newblock In \emph{Thirty-Fourth AAAI Conference on Artificial Intelligence}, 2020.

\bibitem[Bommasani et~al.(2023)Bommasani, Klyman, Longpre, Kapoor, Maslej, Xiong, Zhang, and Liang]{bommasani2023foundationmodeltransparencyindex}
Rishi Bommasani, Kevin Klyman, Shayne Longpre, Sayash Kapoor, Nestor Maslej, Betty Xiong, Daniel Zhang, and Percy Liang.
\newblock The foundation model transparency index, 2023.
\newblock URL \url{https://arxiv.org/abs/2310.12941}.

\bibitem[Broder et~al.(1998)Broder, Charikar, Frieze, and Mitzenmacher]{Broder-1998-minhash}
Andrei~Z. Broder, Moses Charikar, Alan~M. Frieze, and Michael Mitzenmacher.
\newblock Min-wise independent permutations (extended abstract).
\newblock In \emph{Proceedings of the Thirtieth Annual ACM Symposium on Theory of Computing}, STOC '98, pp.\  327–336, New York, NY, USA, 1998. Association for Computing Machinery.
\newblock ISBN 0897919629.
\newblock \doi{10.1145/276698.276781}.
\newblock URL \url{https://doi.org/10.1145/276698.276781}.

\bibitem[Brown et~al.(2020)Brown, Mann, Ryder, Subbiah, Kaplan, Dhariwal, Neelakantan, Shyam, Sastry, Askell, Agarwal, Herbert-Voss, Krueger, Henighan, Child, Ramesh, Ziegler, Wu, Winter, Hesse, Chen, Sigler, Litwin, Gray, Chess, Clark, Berner, McCandlish, Radford, Sutskever, and Amodei]{brown-etal-2020-fewshot}
Tom~B. Brown, Benjamin Mann, Nick Ryder, Melanie Subbiah, Jared Kaplan, Prafulla Dhariwal, Arvind Neelakantan, Pranav Shyam, Girish Sastry, Amanda Askell, Sandhini Agarwal, Ariel Herbert-Voss, Gretchen Krueger, Tom Henighan, Rewon Child, Aditya Ramesh, Daniel~M. Ziegler, Jeffrey Wu, Clemens Winter, Christopher Hesse, Mark Chen, Eric Sigler, Mateusz Litwin, Scott Gray, Benjamin Chess, Jack Clark, Christopher Berner, Sam McCandlish, Alec Radford, Ilya Sutskever, and Dario Amodei.
\newblock Language models are few-shot learners.
\newblock In \emph{Proceedings of the 34th International Conference on Neural Information Processing Systems}, NIPS '20, Red Hook, NY, USA, 2020. Curran Associates Inc.
\newblock ISBN 9781713829546.

\bibitem[Burchell et~al.(2023)Burchell, Birch, Bogoychev, and Heafield]{burchell-etal-2023-open}
Laurie Burchell, Alexandra Birch, Nikolay Bogoychev, and Kenneth Heafield.
\newblock An open dataset and model for language identification.
\newblock In Anna Rogers, Jordan Boyd-Graber, and Naoaki Okazaki (eds.), \emph{Proceedings of the 61st Annual Meeting of the Association for Computational Linguistics (Volume 2: Short Papers)}, pp.\  865--879, Toronto, Canada, July 2023. Association for Computational Linguistics.
\newblock \doi{10.18653/v1/2023.acl-short.75}.
\newblock URL \url{https://aclanthology.org/2023.acl-short.75/}.

\bibitem[Burchell et~al.(2025)Burchell, de~Gibert, Arefyev, Aulamo, Bañón, Chen, Fedorova, Guillou, Haddow, Hajič, Helcl, Henriksson, Klimaszewski, Komulainen, Kutuzov, Kytöniemi, Laippala, Mæhlum, Malik, Mehryary, Mikhailov, Moghe, Myntti, O'Brien, Oepen, Pal, Piha, Pyysalo, Ramírez-Sánchez, Samuel, Stepachev, Tiedemann, Variš, Vojtěchová, and Zaragoza-Bernabeu]{Burchell-HPLT2}
Laurie Burchell, Ona de~Gibert, Nikolay Arefyev, Mikko Aulamo, Marta Bañón, Pinzhen Chen, Mariia Fedorova, Liane Guillou, Barry Haddow, Jan Hajič, Jindřich Helcl, Erik Henriksson, Mateusz Klimaszewski, Ville Komulainen, Andrey Kutuzov, Joona Kytöniemi, Veronika Laippala, Petter Mæhlum, Bhavitvya Malik, Farrokh Mehryary, Vladislav Mikhailov, Nikita Moghe, Amanda Myntti, Dayyán O'Brien, Stephan Oepen, Proyag Pal, Jousia Piha, Sampo Pyysalo, Gema Ramírez-Sánchez, David Samuel, Pavel Stepachev, Jörg Tiedemann, Dušan Variš, Tereza Vojtěchová, and Jaume Zaragoza-Bernabeu.
\newblock An expanded massive multilingual dataset for high-performance language technologies, 2025.
\newblock URL \url{https://arxiv.org/abs/2503.10267}.

\bibitem[Clark et~al.(2018)Clark, Cowhey, Etzioni, Khot, Sabharwal, Schoenick, and Tafjord]{clark-2018-ARC}
Peter Clark, Isaac Cowhey, Oren Etzioni, Tushar Khot, Ashish Sabharwal, Carissa Schoenick, and Oyvind Tafjord.
\newblock Think you have solved question answering? try arc, the ai2 reasoning challenge, 2018.
\newblock URL \url{https://arxiv.org/abs/1803.05457}.

\bibitem[Conneau et~al.(2020)Conneau, Khandelwal, Goyal, Chaudhary, Wenzek, Guzm{\'a}n, Grave, Ott, Zettlemoyer, and Stoyanov]{conneau-etal-xlm-roberta}
Alexis Conneau, Kartikay Khandelwal, Naman Goyal, Vishrav Chaudhary, Guillaume Wenzek, Francisco Guzm{\'a}n, Edouard Grave, Myle Ott, Luke Zettlemoyer, and Veselin Stoyanov.
\newblock Unsupervised cross-lingual representation learning at scale.
\newblock In Dan Jurafsky, Joyce Chai, Natalie Schluter, and Joel Tetreault (eds.), \emph{Proceedings of the 58th Annual Meeting of the Association for Computational Linguistics}, pp.\  8440--8451, Online, July 2020. Association for Computational Linguistics.
\newblock \doi{10.18653/v1/2020.acl-main.747}.
\newblock URL \url{https://aclanthology.org/2020.acl-main.747/}.

\bibitem[de~la Rosa et~al.(2025)de~la Rosa, Mikhailov, Zhang, Wetjen, Samuel, Liu, Braaten, Mæhlum, Birkenes, Kutuzov, Enstad, Farsethås, Brygfjeld, Gulla, Oepen, Velldal, Østgulen, Øvrelid, and Myhre]{delarosa-2025-impactcopyrightedmateriallarge}
Javier de~la Rosa, Vladislav Mikhailov, Lemei Zhang, Freddy Wetjen, David Samuel, Peng Liu, Rolv-Arild Braaten, Petter Mæhlum, Magnus~Breder Birkenes, Andrey Kutuzov, Tita Enstad, Hans~Christian Farsethås, Svein~Arne Brygfjeld, Jon~Atle Gulla, Stephan Oepen, Erik Velldal, Wilfred Østgulen, Liljia Øvrelid, and Aslak~Sira Myhre.
\newblock The impact of copyrighted material on large language models: A norwegian perspective, 2025.
\newblock URL \url{https://arxiv.org/abs/2412.09460}.

\bibitem[Dodge et~al.(2021)Dodge, Sap, Marasovi{\'c}, Agnew, Ilharco, Groeneveld, Mitchell, and Gardner]{dodge-etal-2021-documenting}
Jesse Dodge, Maarten Sap, Ana Marasovi{\'c}, William Agnew, Gabriel Ilharco, Dirk Groeneveld, Margaret Mitchell, and Matt Gardner.
\newblock Documenting large webtext corpora: A case study on the colossal clean crawled corpus.
\newblock In Marie-Francine Moens, Xuanjing Huang, Lucia Specia, and Scott Wen-tau Yih (eds.), \emph{Proceedings of the 2021 Conference on Empirical Methods in Natural Language Processing}, pp.\  1286--1305, Online and Punta Cana, Dominican Republic, November 2021. Association for Computational Linguistics.
\newblock \doi{10.18653/v1/2021.emnlp-main.98}.
\newblock URL \url{https://aclanthology.org/2021.emnlp-main.98/}.

\bibitem[Egbert et~al.(2015)Egbert, Biber, and Davies]{egbert2015}
Jesse Egbert, Douglas Biber, and Mark Davies.
\newblock Developing a user-based method of web register classification.
\newblock \emph{Journal of the Association for Information Science and Technology}, 66, 02 2015.
\newblock \doi{10.1002/asi.23308}.

\bibitem[Fourrier et~al.(2023)Fourrier, Habib, Wolf, and Tunstall]{Fourrier-lighteval}
Clémentine Fourrier, Nathan Habib, Thomas Wolf, and Lewis Tunstall.
\newblock Lighteval: A lightweight framework for llm evaluation, 2023.
\newblock URL \url{https://github.com/huggingface/lighteval}.

\bibitem[Gao et~al.(2020)Gao, Biderman, Black, Golding, Hoppe, Foster, Phang, He, Thite, Nabeshima, et~al.]{gao2020pile}
Leo Gao, Stella Biderman, Sid Black, Laurence Golding, Travis Hoppe, Charles Foster, Jason Phang, Horace He, Anish Thite, Noa Nabeshima, et~al.
\newblock The {P}ile: An 800{GB} dataset of diverse text for language modeling.
\newblock \emph{arXiv preprint arXiv:2101.00027}, 2020.

\bibitem[Gururangan et~al.(2022)Gururangan, Card, Dreier, Gade, Wang, Wang, Zettlemoyer, and Smith]{gururangan-etal-2022-whose}
Suchin Gururangan, Dallas Card, Sarah Dreier, Emily Gade, Leroy Wang, Zeyu Wang, Luke Zettlemoyer, and Noah~A. Smith.
\newblock Whose language counts as high quality? measuring language ideologies in text data selection.
\newblock In Yoav Goldberg, Zornitsa Kozareva, and Yue Zhang (eds.), \emph{Proceedings of the 2022 Conference on Empirical Methods in Natural Language Processing}, pp.\  2562--2580, Abu Dhabi, United Arab Emirates, December 2022. Association for Computational Linguistics.
\newblock \doi{10.18653/v1/2022.emnlp-main.165}.
\newblock URL \url{https://aclanthology.org/2022.emnlp-main.165/}.

\bibitem[He et~al.(2023)He, Gao, and Chen]{he2023debertav3improvingdebertausing}
Pengcheng He, Jianfeng Gao, and Weizhu Chen.
\newblock Debertav3: Improving deberta using electra-style pre-training with gradient-disentangled embedding sharing, 2023.
\newblock URL \url{https://arxiv.org/abs/2111.09543}.

\bibitem[Hendrycks et~al.(2021{\natexlab{a}})Hendrycks, Burns, Basart, Critch, Li, Song, and Steinhardt]{hendrycks2021ethics}
Dan Hendrycks, Collin Burns, Steven Basart, Andrew Critch, Jerry Li, Dawn Song, and Jacob Steinhardt.
\newblock Aligning ai with shared human values.
\newblock \emph{Proceedings of the International Conference on Learning Representations (ICLR)}, 2021{\natexlab{a}}.

\bibitem[Hendrycks et~al.(2021{\natexlab{b}})Hendrycks, Burns, Basart, Zou, Mazeika, Song, and Steinhardt]{hendryckstest2021}
Dan Hendrycks, Collin Burns, Steven Basart, Andy Zou, Mantas Mazeika, Dawn Song, and Jacob Steinhardt.
\newblock Measuring massive multitask language understanding.
\newblock \emph{Proceedings of the International Conference on Learning Representations (ICLR)}, 2021{\natexlab{b}}.

\bibitem[Henriksson et~al.(2024)Henriksson, Myntti, Hellström, Eskelinen, Erten-Johansson, and Laippala]{henriksson2024automaticregisteridentificationopen}
Erik Henriksson, Amanda Myntti, Saara Hellström, Anni Eskelinen, Selcen Erten-Johansson, and Veronika Laippala.
\newblock Automatic register identification for the open web using multilingual deep learning, 2024.
\newblock URL \url{https://arxiv.org/abs/2406.19892}.

\bibitem[Henriksson et~al.(2025)Henriksson, Tarkka, and Ginter]{henriksson2025finerweb10btrefiningwebdata}
Erik Henriksson, Otto Tarkka, and Filip Ginter.
\newblock Finerweb-10bt: Refining web data with llm-based line-level filtering, 2025.
\newblock URL \url{https://arxiv.org/abs/2501.07314}.

\bibitem[Hoffmann et~al.(2022)Hoffmann, Borgeaud, Mensch, Buchatskaya, Cai, Rutherford, de~Las~Casas, Hendricks, Welbl, Clark, Hennigan, Noland, Millican, van~den Driessche, Damoc, Guy, Osindero, Simonyan, Elsen, Vinyals, Rae, and Sifre]{Hoffman-2022-chinchilla}
Jordan Hoffmann, Sebastian Borgeaud, Arthur Mensch, Elena Buchatskaya, Trevor Cai, Eliza Rutherford, Diego de~Las~Casas, Lisa~Anne Hendricks, Johannes Welbl, Aidan Clark, Tom Hennigan, Eric Noland, Katie Millican, George van~den Driessche, Bogdan Damoc, Aurelia Guy, Simon Osindero, Karen Simonyan, Erich Elsen, Oriol Vinyals, Jack~W. Rae, and Laurent Sifre.
\newblock Training compute-optimal large language models.
\newblock In \emph{Proceedings of the 36th International Conference on Neural Information Processing Systems}, NIPS '22, Red Hook, NY, USA, 2022. Curran Associates Inc.
\newblock ISBN 9781713871088.

\bibitem[Kaplan et~al.(2020)Kaplan, McCandlish, Henighan, Brown, Chess, Child, Gray, Radford, Wu, and Amodei]{kaplan2020scalinglawsneurallanguage}
Jared Kaplan, Sam McCandlish, Tom Henighan, Tom~B. Brown, Benjamin Chess, Rewon Child, Scott Gray, Alec Radford, Jeffrey Wu, and Dario Amodei.
\newblock Scaling laws for neural language models, 2020.
\newblock URL \url{https://arxiv.org/abs/2001.08361}.

\bibitem[Kilgarriff \& Grefenstette(2003)Kilgarriff and Grefenstette]{kilgarriff-grefenstette-2003-web-as-corpus}
Adam Kilgarriff and Gregory Grefenstette.
\newblock Introduction to the special issue on the web as corpus.
\newblock \emph{Computational Linguistics}, 29\penalty0 (3):\penalty0 333--348, 2003.
\newblock \doi{10.1162/089120103322711569}.
\newblock URL \url{https://aclanthology.org/J03-3001/}.

\bibitem[Kuzman et~al.(2023{\natexlab{a}})Kuzman, Mozetič, and Ljubešić]{Kuzman-genre-identification-2023}
Taja Kuzman, Igor Mozetič, and Nikola Ljubešić.
\newblock Automatic genre identification for robust enrichment of massive text collections: Investigation of classification methods in the era of large language models.
\newblock \emph{Machine Learning and Knowledge Extraction}, 5\penalty0 (3):\penalty0 1149–1175, Sep 2023{\natexlab{a}}.
\newblock ISSN 2504-4990.
\newblock \doi{10.3390/make5030059}.
\newblock URL \url{http://dx.doi.org/10.3390/make5030059}.

\bibitem[Kuzman et~al.(2023{\natexlab{b}})Kuzman, Rupnik, and Ljube{\v{s}}i{\'c}]{kuzman-etal-2023-get}
Taja Kuzman, Peter Rupnik, and Nikola Ljube{\v{s}}i{\'c}.
\newblock Get to know your parallel data: Performing {E}nglish variety and genre classification over {M}a{C}o{C}u corpora.
\newblock In Yves Scherrer, Tommi Jauhiainen, Nikola Ljube{\v{s}}i{\'c}, Preslav Nakov, J{\"o}rg Tiedemann, and Marcos Zampieri (eds.), \emph{Tenth Workshop on NLP for Similar Languages, Varieties and Dialects (VarDial 2023)}, pp.\  91--103, Dubrovnik, Croatia, May 2023{\natexlab{b}}. Association for Computational Linguistics.
\newblock \doi{10.18653/v1/2023.vardial-1.9}.
\newblock URL \url{https://aclanthology.org/2023.vardial-1.9/}.

\bibitem[Laippala et~al.(2023)Laippala, R{\"o}nnqvist, Oinonen, Kyr{\"o}l{\"a}inen, Salmela, Biber, Egbert, and Pyysalo]{laippala_ronnqvist2023}
Veronika Laippala, Samuel R{\"o}nnqvist, Miika Oinonen, {Aki Juhani} Kyr{\"o}l{\"a}inen, Anna Salmela, Douglas Biber, Jesse Egbert, and Sampo Pyysalo.
\newblock Register identification from the unrestricted open web using the corpus of online registers of english.
\newblock \emph{Language Resources and Evaluation}, 57\penalty0 (3):\penalty0 1045--1079, September 2023.
\newblock ISSN 1574-020X.
\newblock \doi{10.1007/s10579-022-09624-1}.
\newblock Funding Information: Open Access funding provided by University of Turku (UTU) including Turku University Central Hospital. Funding was provided by Academy of Finland (Grant No. 331297), Emil Aaltosen s{\"a}{\"a}ti{\"o}, National Science Foundation (Grant No. 1147581). Publisher Copyright: {\textcopyright} 2022, The Author(s).

\bibitem[Lee et~al.(2022)Lee, Ippolito, Nystrom, Zhang, Eck, Callison-Burch, and Carlini]{lee-etal-2022-deduplicating}
Katherine Lee, Daphne Ippolito, Andrew Nystrom, Chiyuan Zhang, Douglas Eck, Chris Callison-Burch, and Nicholas Carlini.
\newblock Deduplicating training data makes language models better.
\newblock In Smaranda Muresan, Preslav Nakov, and Aline Villavicencio (eds.), \emph{Proceedings of the 60th Annual Meeting of the Association for Computational Linguistics (Volume 1: Long Papers)}, pp.\  8424--8445, Dublin, Ireland, May 2022. Association for Computational Linguistics.
\newblock \doi{10.18653/v1/2022.acl-long.577}.
\newblock URL \url{https://aclanthology.org/2022.acl-long.577/}.

\bibitem[Li et~al.(2024)Li, Fang, Smyrnis, Ivgi, Jordan, Gadre, Bansal, Guha, Keh, Arora, Garg, Xin, Muennighoff, Heckel, Mercat, Chen, Gururangan, Wortsman, Albalak, Bitton, Nezhurina, Abbas, Hsieh, Ghosh, Gardner, Kilian, Zhang, Shao, Pratt, Sanyal, Ilharco, Daras, Marathe, Gokaslan, Zhang, Chandu, Nguyen, Vasiljevic, Kakade, Song, Sanghavi, Faghri, Oh, Zettlemoyer, Lo, El-Nouby, Pouransari, Toshev, Wang, Groeneveld, Soldaini, Koh, Jitsev, Kollar, Dimakis, Carmon, Dave, Schmidt, and Shankar]{li2024datacomplmsearchgenerationtraining}
Jeffrey Li, Alex Fang, Georgios Smyrnis, Maor Ivgi, Matt Jordan, Samir Gadre, Hritik Bansal, Etash Guha, Sedrick Keh, Kushal Arora, Saurabh Garg, Rui Xin, Niklas Muennighoff, Reinhard Heckel, Jean Mercat, Mayee Chen, Suchin Gururangan, Mitchell Wortsman, Alon Albalak, Yonatan Bitton, Marianna Nezhurina, Amro Abbas, Cheng-Yu Hsieh, Dhruba Ghosh, Josh Gardner, Maciej Kilian, Hanlin Zhang, Rulin Shao, Sarah Pratt, Sunny Sanyal, Gabriel Ilharco, Giannis Daras, Kalyani Marathe, Aaron Gokaslan, Jieyu Zhang, Khyathi Chandu, Thao Nguyen, Igor Vasiljevic, Sham Kakade, Shuran Song, Sujay Sanghavi, Fartash Faghri, Sewoong Oh, Luke Zettlemoyer, Kyle Lo, Alaaeldin El-Nouby, Hadi Pouransari, Alexander Toshev, Stephanie Wang, Dirk Groeneveld, Luca Soldaini, Pang~Wei Koh, Jenia Jitsev, Thomas Kollar, Alexandros~G. Dimakis, Yair Carmon, Achal Dave, Ludwig Schmidt, and Vaishaal Shankar.
\newblock Datacomp-lm: In search of the next generation of training sets for language models, 2024.
\newblock URL \url{https://arxiv.org/abs/2406.11794}.

\bibitem[Longpre et~al.(2024)Longpre, Yauney, Reif, Lee, Roberts, Zoph, Zhou, Wei, Robinson, Mimno, and Ippolito]{longpre-etal-2024-pretrainers}
Shayne Longpre, Gregory Yauney, Emily Reif, Katherine Lee, Adam Roberts, Barret Zoph, Denny Zhou, Jason Wei, Kevin Robinson, David Mimno, and Daphne Ippolito.
\newblock A pretrainer`s guide to training data: Measuring the effects of data age, domain coverage, quality, {\&} toxicity.
\newblock In Kevin Duh, Helena Gomez, and Steven Bethard (eds.), \emph{Proceedings of the 2024 Conference of the North American Chapter of the Association for Computational Linguistics: Human Language Technologies (Volume 1: Long Papers)}, pp.\  3245--3276, Mexico City, Mexico, June 2024. Association for Computational Linguistics.
\newblock \doi{10.18653/v1/2024.naacl-long.179}.
\newblock URL \url{https://aclanthology.org/2024.naacl-long.179/}.

\bibitem[Mihaylov et~al.(2018)Mihaylov, Clark, Khot, and Sabharwal]{OpenBookQA2018}
Todor Mihaylov, Peter Clark, Tushar Khot, and Ashish Sabharwal.
\newblock Can a suit of armor conduct electricity? a new dataset for open book question answering.
\newblock In \emph{EMNLP}, 2018.

\bibitem[Muennighoff et~al.(2023)Muennighoff, Rush, Barak, Le~Scao, Piktus, Tazi, Pyysalo, Wolf, and Raffel]{Muenninghoff-2023-scaling}
Niklas Muennighoff, Alexander~M. Rush, Boaz Barak, Teven Le~Scao, Aleksandra Piktus, Nouamane Tazi, Sampo Pyysalo, Thomas Wolf, and Colin Raffel.
\newblock Scaling data-constrained language models.
\newblock In \emph{Proceedings of the 37th International Conference on Neural Information Processing Systems}, NIPS '23, Red Hook, NY, USA, 2023. Curran Associates Inc.

\bibitem[Myntti et~al.(2024)Myntti, Repo, Freyermuth, Kanner, Laippala, and Henriksson]{myntti-etal-2024-intersecting}
Amanda Myntti, Liina Repo, Elian Freyermuth, Antti Kanner, Veronika Laippala, and Erik Henriksson.
\newblock Intersecting register and genre: Understanding the contents of web-crawled corpora.
\newblock In Mika H{\"a}m{\"a}l{\"a}inen, Emily {\"O}hman, So~Miyagawa, Khalid Alnajjar, and Yuri Bizzoni (eds.), \emph{Proceedings of the 4th International Conference on Natural Language Processing for Digital Humanities}, pp.\  386--397, Miami, USA, November 2024. Association for Computational Linguistics.
\newblock \doi{10.18653/v1/2024.nlp4dh-1.38}.
\newblock URL \url{https://aclanthology.org/2024.nlp4dh-1.38/}.

\bibitem[Ostendorff et~al.(2024)Ostendorff, Suarez, Lage, and Rehm]{ostendorff2024llmdatasets}
Malte Ostendorff, Pedro~Ortiz Suarez, Lucas~Fonseca Lage, and Georg Rehm.
\newblock {LLM}-datasets: An open framework for pretraining datasets of large language models.
\newblock In \emph{First Conference on Language Modeling}, 2024.
\newblock URL \url{https://openreview.net/forum?id=5RdIMlGLXL}.

\bibitem[Penedo et~al.(2023)Penedo, Malartic, Hesslow, Cojocaru, Alobeidli, Cappelli, Pannier, Almazrouei, and Launay]{penedo-2023-refinedweb}
Guilherme Penedo, Quentin Malartic, Daniel Hesslow, Ruxandra Cojocaru, Hamza Alobeidli, Alessandro Cappelli, Baptiste Pannier, Ebtesam Almazrouei, and Julien Launay.
\newblock The refinedweb dataset for falcon llm: outperforming curated corpora with web data only.
\newblock In \emph{Proceedings of the 37th International Conference on Neural Information Processing Systems}, NIPS '23, Red Hook, NY, USA, 2023. Curran Associates Inc.

\bibitem[Penedo et~al.(2024{\natexlab{a}})Penedo, Kydl\'{\i}\v{c}ek, allal, Lozhkov, Mitchell, Raffel, Von~Werra, and Wolf]{penedo2024finewebdatasetsdecantingweb}
Guilherme Penedo, Hynek Kydl\'{\i}\v{c}ek, Loubna~Ben allal, Anton Lozhkov, Margaret Mitchell, Colin Raffel, Leandro Von~Werra, and Thomas Wolf.
\newblock The fineweb datasets: Decanting the web for the finest text data at scale.
\newblock In A.~Globerson, L.~Mackey, D.~Belgrave, A.~Fan, U.~Paquet, J.~Tomczak, and C.~Zhang (eds.), \emph{Advances in Neural Information Processing Systems}, volume~37, pp.\  30811--30849. Curran Associates, Inc., 2024{\natexlab{a}}.
\newblock URL \url{https://proceedings.neurips.cc/paper_files/paper/2024/file/370df50ccfdf8bde18f8f9c2d9151bda-Paper-Datasets_and_Benchmarks_Track.pdf}.

\bibitem[Penedo et~al.(2024{\natexlab{b}})Penedo, Kydlíček, Sabolčec, Messmer, Foroutan, Jaggi, von Werra, and Wolf]{penedo2024fineweb-2}
Guilherme Penedo, Hynek Kydlíček, Vinko Sabolčec, Bettina Messmer, Negar Foroutan, Martin Jaggi, Leandro von Werra, and Thomas Wolf.
\newblock Fineweb2: A sparkling update with 1000s of languages, December 2024{\natexlab{b}}.
\newblock URL \url{https://huggingface.co/datasets/HuggingFaceFW/fineweb-2}.

\bibitem[Petrenz \& Webber(2011)Petrenz and Webber]{petrenz-webber-2011-squibs}
Philipp Petrenz and Bonnie Webber.
\newblock {S}quibs: Stable classification of text genres.
\newblock \emph{Computational Linguistics}, 37\penalty0 (2):\penalty0 385--393, June 2011.
\newblock \doi{10.1162/COLI_a_00052}.
\newblock URL \url{https://aclanthology.org/J11-2004/}.

\bibitem[Radford et~al.(2019)Radford, Wu, Child, Luan, Amodei, and Sutskever]{radford2019gpt2}
Alec Radford, Jeffrey Wu, Rewon Child, David Luan, Dario Amodei, and Ilya Sutskever.
\newblock Language models are unsupervised multitask learners.
\newblock \emph{OpenAI}, 2019.
\newblock URL \url{https://cdn.openai.com/better-language-models/language_models_are_unsupervised_multitask_learners.pdf}.
\newblock Accessed: 2024-11-15.

\bibitem[Rae et~al.(2022)Rae, Borgeaud, Cai, Millican, Hoffmann, Song, Aslanides, Henderson, Ring, Young, Rutherford, Hennigan, Menick, Cassirer, Powell, van~den Driessche, Hendricks, Rauh, Huang, Glaese, Welbl, Dathathri, Huang, Uesato, Mellor, Higgins, Creswell, McAleese, Wu, Elsen, Jayakumar, Buchatskaya, Budden, Sutherland, Simonyan, Paganini, Sifre, Martens, Li, Kuncoro, Nematzadeh, Gribovskaya, Donato, Lazaridou, Mensch, Lespiau, Tsimpoukelli, Grigorev, Fritz, Sottiaux, Pajarskas, Pohlen, Gong, Toyama, de~Masson~d'Autume, Li, Terzi, Mikulik, Babuschkin, Clark, de~Las~Casas, Guy, Jones, Bradbury, Johnson, Hechtman, Weidinger, Gabriel, Isaac, Lockhart, Osindero, Rimell, Dyer, Vinyals, Ayoub, Stanway, Bennett, Hassabis, Kavukcuoglu, and Irving]{rae2022scalinglanguagemodelsmethods}
Jack~W. Rae, Sebastian Borgeaud, Trevor Cai, Katie Millican, Jordan Hoffmann, Francis Song, John Aslanides, Sarah Henderson, Roman Ring, Susannah Young, Eliza Rutherford, Tom Hennigan, Jacob Menick, Albin Cassirer, Richard Powell, George van~den Driessche, Lisa~Anne Hendricks, Maribeth Rauh, Po-Sen Huang, Amelia Glaese, Johannes Welbl, Sumanth Dathathri, Saffron Huang, Jonathan Uesato, John Mellor, Irina Higgins, Antonia Creswell, Nat McAleese, Amy Wu, Erich Elsen, Siddhant Jayakumar, Elena Buchatskaya, David Budden, Esme Sutherland, Karen Simonyan, Michela Paganini, Laurent Sifre, Lena Martens, Xiang~Lorraine Li, Adhiguna Kuncoro, Aida Nematzadeh, Elena Gribovskaya, Domenic Donato, Angeliki Lazaridou, Arthur Mensch, Jean-Baptiste Lespiau, Maria Tsimpoukelli, Nikolai Grigorev, Doug Fritz, Thibault Sottiaux, Mantas Pajarskas, Toby Pohlen, Zhitao Gong, Daniel Toyama, Cyprien de~Masson~d'Autume, Yujia Li, Tayfun Terzi, Vladimir Mikulik, Igor Babuschkin, Aidan Clark, Diego de~Las~Casas, Aurelia Guy, Chris Jones,
  James Bradbury, Matthew Johnson, Blake Hechtman, Laura Weidinger, Iason Gabriel, William Isaac, Ed~Lockhart, Simon Osindero, Laura Rimell, Chris Dyer, Oriol Vinyals, Kareem Ayoub, Jeff Stanway, Lorrayne Bennett, Demis Hassabis, Koray Kavukcuoglu, and Geoffrey Irving.
\newblock Scaling language models: Methods, analysis \& insights from training gopher, 2022.
\newblock URL \url{https://arxiv.org/abs/2112.11446}.

\bibitem[Raffel et~al.(2020)Raffel, Shazeer, Roberts, Lee, Narang, Matena, Zhou, Li, and Liu]{raffel-2020-transferlearning}
Colin Raffel, Noam Shazeer, Adam Roberts, Katherine Lee, Sharan Narang, Michael Matena, Yanqi Zhou, Wei Li, and Peter~J. Liu.
\newblock Exploring the limits of transfer learning with a unified text-to-text transformer.
\newblock \emph{J. Mach. Learn. Res.}, 21\penalty0 (1), January 2020.
\newblock ISSN 1532-4435.

\bibitem[Repo et~al.(2021)Repo, Skantsi, R{\"o}nnqvist, Hellstr{\"o}m, Oinonen, Salmela, Biber, Egbert, Pyysalo, and Laippala]{repo2021zeroshot}
Liina Repo, Valtteri Skantsi, Samuel R{\"o}nnqvist, Saara Hellstr{\"o}m, Miika Oinonen, Anna Salmela, Douglas Biber, Jesse Egbert, Sampo Pyysalo, and Veronika Laippala.
\newblock Beyond the english web: Zero-shot cross-lingual and lightweight monolingual classification of registers.
\newblock In \emph{Proceedings of the 16th Conference of the European Chapter of the Association for Computational Linguistics: Student Research Workshop}, 2021.
\newblock URL \url{https://aclanthology.org/2021.eacl-srw.24.pdf}.

\bibitem[R{\"o}nnqvist et~al.(2021)R{\"o}nnqvist, Skantsi, Oinonen, and Laippala]{ronnqvist-etal-2021-multilingual}
Samuel R{\"o}nnqvist, Valtteri Skantsi, Miika Oinonen, and Veronika Laippala.
\newblock Multilingual and zero-shot is closing in on monolingual web register classification.
\newblock In \emph{Proceedings of the 23rd Nordic Conference on Computational Linguistics (NoDaLiDa)}, pp.\  157--165, Reykjavik, Iceland (Online), 2021. Link{\"o}ping University Electronic Press, Sweden.
\newblock URL \url{https://aclanthology.org/2021.nodalida-main.16}.

\bibitem[Sachdeva et~al.(2024)Sachdeva, Coleman, Kang, Ni, Hong, Chi, Caverlee, McAuley, and Cheng]{sachdeva2024traindataefficientllms}
Noveen Sachdeva, Benjamin Coleman, Wang-Cheng Kang, Jianmo Ni, Lichan Hong, Ed~H. Chi, James Caverlee, Julian McAuley, and Derek~Zhiyuan Cheng.
\newblock How to train data-efficient llms, 2024.
\newblock URL \url{https://arxiv.org/abs/2402.09668}.

\bibitem[Sakaguchi et~al.(2021)Sakaguchi, Bras, Bhagavatula, and Choi]{sakaguchi-2021-winogrande}
Keisuke Sakaguchi, Ronan~Le Bras, Chandra Bhagavatula, and Yejin Choi.
\newblock Winogrande: an adversarial winograd schema challenge at scale.
\newblock \emph{Commun. ACM}, 64\penalty0 (9):\penalty0 99–106, August 2021.
\newblock ISSN 0001-0782.
\newblock \doi{10.1145/3474381}.
\newblock URL \url{https://doi.org/10.1145/3474381}.

\bibitem[Santini(2007)]{santini-2007-automatic-web-genre}
Marina Santini.
\newblock \emph{Automatic identification of genre in web pages}.
\newblock PhD thesis, University of Brighton, 2007.
\newblock Unpublished doctoral dissertation.

\bibitem[Sap et~al.(2019)Sap, Rashkin, Chen, Le~Bras, and Choi]{sap-etal-2019-social}
Maarten Sap, Hannah Rashkin, Derek Chen, Ronan Le~Bras, and Yejin Choi.
\newblock Social {IQ}a: Commonsense reasoning about social interactions.
\newblock In Kentaro Inui, Jing Jiang, Vincent Ng, and Xiaojun Wan (eds.), \emph{Proceedings of the 2019 Conference on Empirical Methods in Natural Language Processing and the 9th International Joint Conference on Natural Language Processing (EMNLP-IJCNLP)}, pp.\  4463--4473, Hong Kong, China, November 2019. Association for Computational Linguistics.
\newblock \doi{10.18653/v1/D19-1454}.
\newblock URL \url{https://aclanthology.org/D19-1454/}.

\bibitem[Sharoff(2020)]{sharoff-2020-know}
Serge Sharoff.
\newblock Know thy corpus! robust methods for digital curation of web corpora.
\newblock In Nicoletta Calzolari, Fr{\'e}d{\'e}ric B{\'e}chet, Philippe Blache, Khalid Choukri, Christopher Cieri, Thierry Declerck, Sara Goggi, Hitoshi Isahara, Bente Maegaard, Joseph Mariani, H{\'e}l{\`e}ne Mazo, Asuncion Moreno, Jan Odijk, and Stelios Piperidis (eds.), \emph{Proceedings of the Twelfth Language Resources and Evaluation Conference}, pp.\  2453--2460, Marseille, France, May 2020. European Language Resources Association.
\newblock ISBN 979-10-95546-34-4.
\newblock URL \url{https://aclanthology.org/2020.lrec-1.298/}.

\bibitem[Sharoff et~al.(2010)Sharoff, Wu, and Markert]{sharoff-etal-2010-web-genre}
Serge Sharoff, Zhili Wu, and Katja Markert.
\newblock The web library of babel: evaluating genre collections.
\newblock In Nicoletta Calzolari, Khalid Choukri, Bente Maegaard, Joseph Mariani, Jan Odijk, Stelios Piperidis, Mike Rosner, and Daniel Tapias (eds.), \emph{Proceedings of the Seventh International Conference on Language Resources and Evaluation ({LREC}`10)}, Valletta, Malta, May 2010. European Language Resources Association (ELRA).
\newblock URL \url{https://aclanthology.org/L10-1011/}.

\bibitem[Su et~al.(2024)Su, Kong, Lin, Jennings, Norick, Kliegl, Patwary, Shoeybi, and Catanzaro]{su2024nemotroncctransformingcommoncrawl}
Dan Su, Kezhi Kong, Ying Lin, Joseph Jennings, Brandon Norick, Markus Kliegl, Mostofa Patwary, Mohammad Shoeybi, and Bryan Catanzaro.
\newblock Nemotron-cc: Transforming common crawl into a refined long-horizon pretraining dataset, 2024.
\newblock URL \url{https://arxiv.org/abs/2412.02595}.

\bibitem[Talmor et~al.(2019)Talmor, Herzig, Lourie, and Berant]{talmor-etal-2019-commonsenseqa}
Alon Talmor, Jonathan Herzig, Nicholas Lourie, and Jonathan Berant.
\newblock {C}ommonsense{QA}: A question answering challenge targeting commonsense knowledge.
\newblock In Jill Burstein, Christy Doran, and Thamar Solorio (eds.), \emph{Proceedings of the 2019 Conference of the North {A}merican Chapter of the Association for Computational Linguistics: Human Language Technologies, Volume 1 (Long and Short Papers)}, pp.\  4149--4158, Minneapolis, Minnesota, June 2019. Association for Computational Linguistics.
\newblock \doi{10.18653/v1/N19-1421}.
\newblock URL \url{https://aclanthology.org/N19-1421/}.

\bibitem[Taylor et~al.(2022)Taylor, Kardas, Cucurull, Scialom, Hartshorn, Saravia, Poulton, Kerkez, and Stojnic]{taylor2022galacticalargelanguagemodel}
Ross Taylor, Marcin Kardas, Guillem Cucurull, Thomas Scialom, Anthony Hartshorn, Elvis Saravia, Andrew Poulton, Viktor Kerkez, and Robert Stojnic.
\newblock Galactica: A large language model for science, 2022.
\newblock URL \url{https://arxiv.org/abs/2211.09085}.

\bibitem[Tirumala et~al.(2023)Tirumala, Simig, Aghajanyan, and Morcos]{tirumala-etal-2023-improving-pretraining}
Kushal Tirumala, Daniel Simig, Armen Aghajanyan, and Ari Morcos.
\newblock D4: Improving llm pretraining via document de-duplication and diversification.
\newblock In A.~Oh, T.~Naumann, A.~Globerson, K.~Saenko, M.~Hardt, and S.~Levine (eds.), \emph{Advances in Neural Information Processing Systems}, volume~36, pp.\  53983--53995. Curran Associates, Inc., 2023.
\newblock URL \url{https://proceedings.neurips.cc/paper_files/paper/2023/file/a8f8cbd7f7a5fb2c837e578c75e5b615-Paper-Datasets_and_Benchmarks.pdf}.

\bibitem[Villalobos et~al.(2024)Villalobos, Ho, Sevilla, Besiroglu, Heim, and Hobbhahn]{Villalobos-etal-2024-run-out-of-data}
Pablo Villalobos, Anson Ho, Jaime Sevilla, Tamay Besiroglu, Lennart Heim, and Marius Hobbhahn.
\newblock Position: will we run out of data? limits of llm scaling based on human-generated data.
\newblock In \emph{Proceedings of the 41st International Conference on Machine Learning}, ICML'24. JMLR.org, 2024.

\bibitem[Weber et~al.(2024)Weber, Fu, Anthony, Oren, Adams, Alexandrov, Lyu, Nguyen, Yao, Adams, Athiwaratkun, Chalamala, Chen, Ryabinin, Dao, Liang, Re, Rish, and Zhang]{weber2024redpajama}
Maurice Weber, Daniel~Y Fu, Quentin~Gregory Anthony, Yonatan Oren, Shane Adams, Anton Alexandrov, Xiaozhong Lyu, Huu Nguyen, Xiaozhe Yao, Virginia Adams, Ben Athiwaratkun, Rahul Chalamala, Kezhen Chen, Max Ryabinin, Tri Dao, Percy Liang, Christopher Re, Irina Rish, and Ce~Zhang.
\newblock Redpajama: an open dataset for training large language models.
\newblock In \emph{The Thirty-eight Conference on Neural Information Processing Systems Datasets and Benchmarks Track}, 2024.
\newblock URL \url{https://openreview.net/forum?id=lnuXaRpwvw}.

\bibitem[Xue et~al.(2023)Xue, Fu, Zhou, Zheng, and You]{Xue-2023-insights-from-scaling}
Fuzhao Xue, Yao Fu, Wangchunshu Zhou, Zangwei Zheng, and Yang You.
\newblock To repeat or not to repeat: Insights from scaling llm under token-crisis.
\newblock In A.~Oh, T.~Naumann, A.~Globerson, K.~Saenko, M.~Hardt, and S.~Levine (eds.), \emph{Advances in Neural Information Processing Systems}, volume~36, pp.\  59304--59322. Curran Associates, Inc., 2023.
\newblock URL \url{https://proceedings.neurips.cc/paper_files/paper/2023/file/b9e472cd579c83e2f6aa3459f46aac28-Paper-Conference.pdf}.

\bibitem[Zellers et~al.(2019)Zellers, Holtzman, Bisk, Farhadi, and Choi]{zellers2019hellaswag}
Rowan Zellers, Ari Holtzman, Yonatan Bisk, Ali Farhadi, and Yejin Choi.
\newblock Hellaswag: Can a machine really finish your sentence?
\newblock In \emph{Proceedings of the 57th Annual Meeting of the Association for Computational Linguistics}, 2019.

\end{thebibliography}
\bibliographystyle{colm2025_conference}

\appendix
\newpage

\section{Examples of texts in register classes}
\label{sec:appendix-text-examples}

Below we show examples of texts classified as selected registers. These texts were chosen randomly but with a focus towards brevity (with some texts additionally truncated), general topics, and non-toxic language use. We also omit new-line characters.

\HIa
\begin{quote}
    Home \& DIY How To Cut Your Own Hair At Home 
    With countless non-essential businesses closing due to the government-implemented lockdowns, many people have been forced to be creative and resourceful in order to get by. Case in point, not everyone of us are lucky enough to be quarantined at home with a professional hairdresser. So we do the best we can -- which often means cutting our own hair. [...]
\end{quote}

\IDa 
\begin{quote}
    I tried last night to compile a standalone which I've compiled approximately 1, 000, 000 times before, and have been met with a long series of very lengthy crash reports. As far as I can tell, something in the compiler is trying to install every framework on my computer, and I sometimes wind up with 1 gig+ .mxf files. Naturally, little dinky tests compile fine, and I can compile all of the modules that make up my patch, but not the whole thing at once. I did a super- clean Max reinstall, no third-party objects aside a couple of Lobjects which are up to date, and I'm stumped. The only difference I can think of since I compiled this a few weeks ago is an upgrade to 4.5. Anyone have any light to throw on this? Im happy to send crashlogs, but they're big. 10.5.2, Max 4.6, Jitter 1.6.3 thanks, M
\end{quote}

\MTa 
\begin{quote}
    Incredible Interior Design Sketches Good Looking Interior Designer Remarkable Interior Design Exceptional Interior Design Sketches Cozy Interior Design Sketches Chic Interior Design Sketches Cool Interior Design Sketches
\end{quote}

\NEa
\begin{quote}
    Wal-Mart has signed a storm water settlement agreement with the Connecticut Department of Environmental Protection concerning alleged violations at 20 Wal-Mart stores and two SAM'S CLUB locations in the state. Under the agreement, Wal-Mart will pay \$600,000 in civil penalties for violations alleged to have taken place between 1996 and 2003. Wal-Mart also will contribute \$550,000 to two different supplemental environmental projects--\$500,000 to assist municipalities in addressing storm water issues, and \$50,000 for environmental projects in the Connecticut River Watershed. [...]
\end{quote}

\OPa 
\begin{quote}
    Once again a very nice stop for the night. Staff has been very well trained, all are friendly and helpful with one exception/the young man who checked us in. Breakfast was so appreciated after 12 nights at a Holiday Inn Express which was ok but does not have your choices. Lynne does an especially good job/appreciated all her help. They... More
\end{quote}

\section{Model training setup}
\label{sec:appendix-training-setup}

Models were trained following the setup of \citet{penedo2024finewebdatasetsdecantingweb}.  We used the GPT-2 tokeniser, Llama architecture and the same training settings: 1.71B parameters (1.82B with embeddings), sequence length of 2048 tokens, and a global batch size of $\sim$2 million tokens. Models were trained for 50,000 iterations, which amounts to 100 billion tokens. Training was done on \LUMI supercomputer on 16 nodes, each with a single 64-core CPU and 4 MI250x GPUs with dual-GCD (graphics compute die). We used NVIDIA's \texttt{Megatron-LM} \footnote{\url{https://github.com/NVIDIA/Megatron-LM}} training framework instead of HuggingFace's \texttt{nanotron}\footnote{\url{https://github.com/huggingface/nanotron}} framework used by \citet{penedo2024finewebdatasetsdecantingweb}. 

Training each model took approximately 84 hours, amounting to 10,752 GPU hours per model with the node setup described above, with average performance of 25 teraflops per second. Evaluation was carried out on a single GPU, and with 19 models, each with 50 checkpoints requiring approximately 20 minutes to evaluate, this added 300 GPU hours to the total computational cost. Other processing tasks, such as sampling and tokenisation, required no GPU resources and used a comparatively negligible amount of CPU hours.

\section{Comparison to other data curation schemes}
\label{sec:appendix-comparison}

As discussed in Section \ref{sec:Related-work}, many different training data selection schemes and tools have been introduced in the literature. To quantify the connection and possible similarities between our register scheme and three other publicly available quality classification tools, FineWeb-edu classifier \citep{penedo2024finewebdatasetsdecantingweb}, the DCLM quality classifier \citep{li2024datacomplmsearchgenerationtraining}, and NVIDIA's NemoCurator Quality Classifier DeBERTa \citep{he2023debertav3improvingdebertausing}, we classify a sample of over 50 000 documents from our pretraining material with all four classifiers. We compare the selected tool to our register classification scheme by conducting a $\chi^2$-test to evaluate dependence and also report metrics pertaining to the possible overlap.

The FineWeb-edu classifier assigns documents an ordinal label ranging from 0 to 5, the NemoCurator a label 0, 1, or 2, and the DCLM classifier uses a binary label (0,1). For registers, we use the classes in Table~\ref{tab:register-descriptions} under \quot{Individual registers}. Obtained results from a $\chi^2$-test (considering FineWeb-edu and NemoCurator's labels nominal) show that all quality classification tools are dependent on the register classification results with $p<0.01$. This dependence between the classification schemes is to be expected, as all schemes are used as a proxy for \quot{quality} in LLM pretraining. To evaluate the strength of this connection, we calculated Cramer's V, which yielded the following results:

\begin{compactitem}
    \item Registers -- DCLM classifier: 0.136
    \item Registers -- FineWeb-edu classifier: 0.221
    \item Registers -- NemoCurator classifier: 0.419
\end{compactitem}
In the context of machine learning, these values indicate a low to moderate connection between the classifiers, with the NemoCurator showing the greatest association, which can be considered somewhat substantial. To account for the two ordinal classification schemes, FineWeb-edu and NemoCurator, we also calculated the $\eta^2$-metric, which measures the amount of impact a variable has on the variance of another. These results were 0.173 and 0.273, for FineWeb-edu and NemoCurator, respectively, which similarly show a strong connection between the schemes but do not imply direct overlap. 

\begin{figure}
    \centering
    \includegraphics[width=0.7\linewidth]{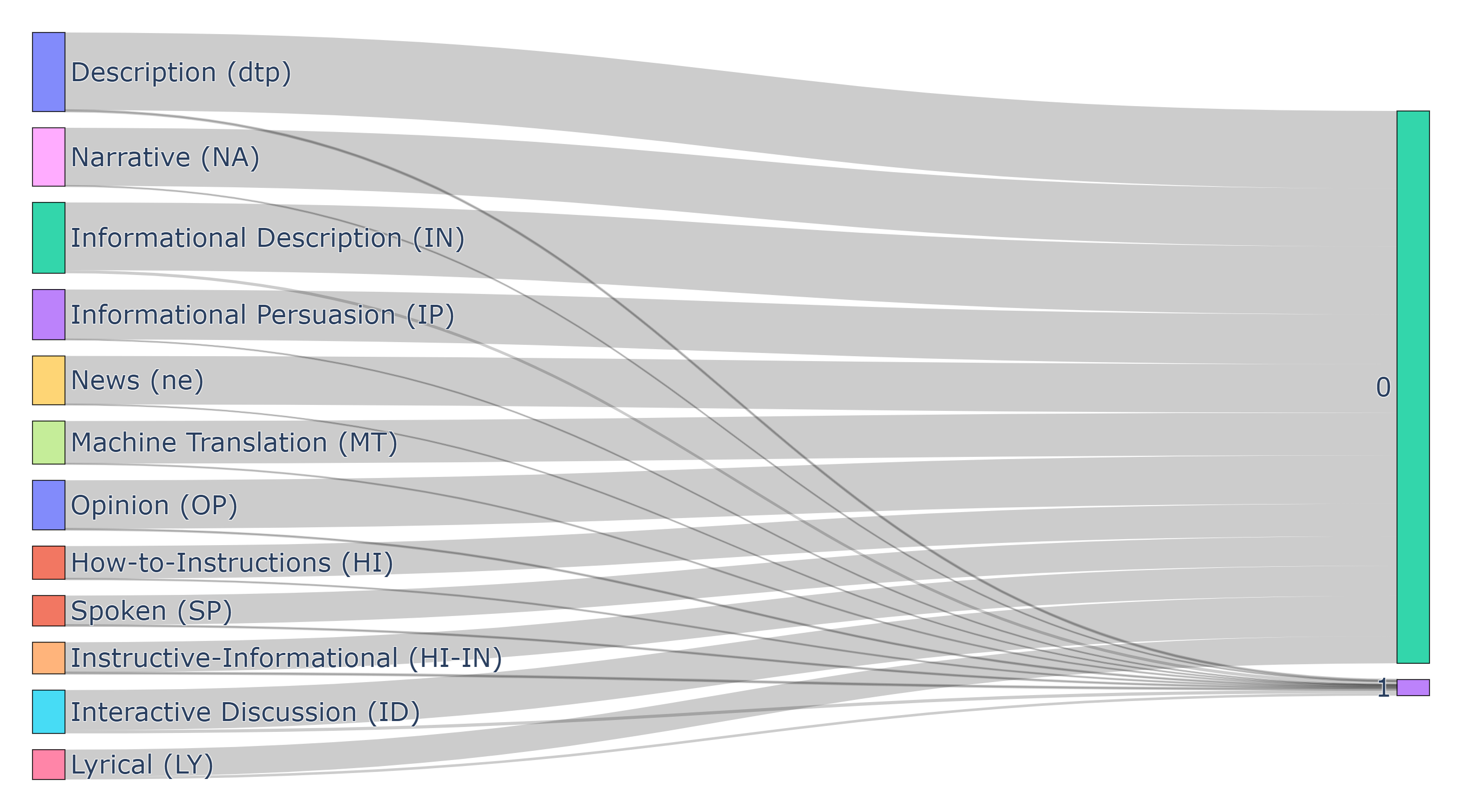}
    \caption{Visualised cross-classification for our register scheme and the DCLM quality classifier. DCLM label 0 corresponds to low quality and label 1 to high quality.}
    \label{fig:sankey-dclm}
\end{figure}

\begin{figure}
    \centering
    \includegraphics[width=0.7\linewidth]{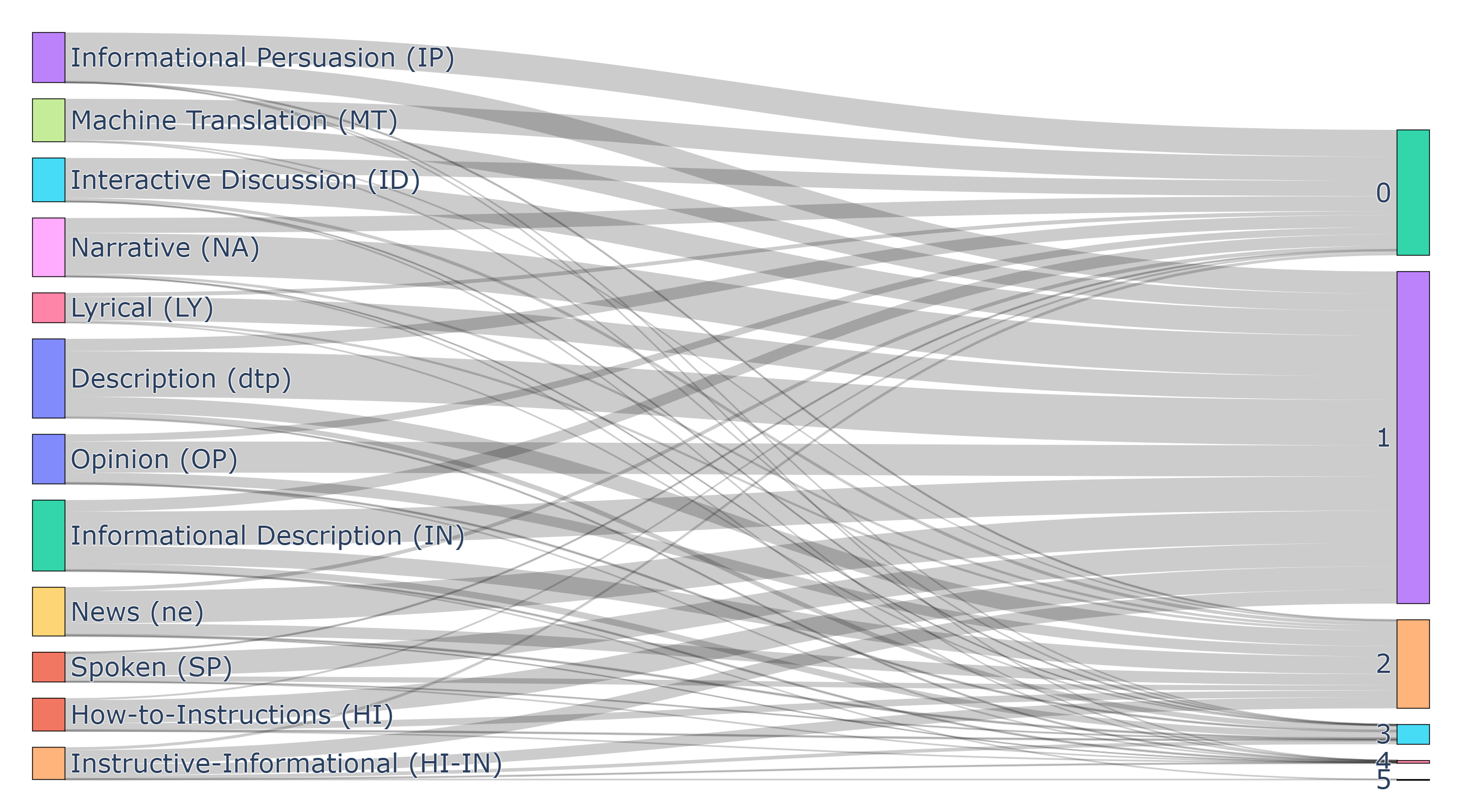}
    \caption{Visualised cross-classification for our register scheme and the FineWeb-edu classifier. FineWeb-edu label 0 corresponds to the lowest quality and label 5 to the highest quality.}
    \label{fig:sankey-edu}
\end{figure}

\begin{figure}
    \centering
    \includegraphics[width=0.7\linewidth]{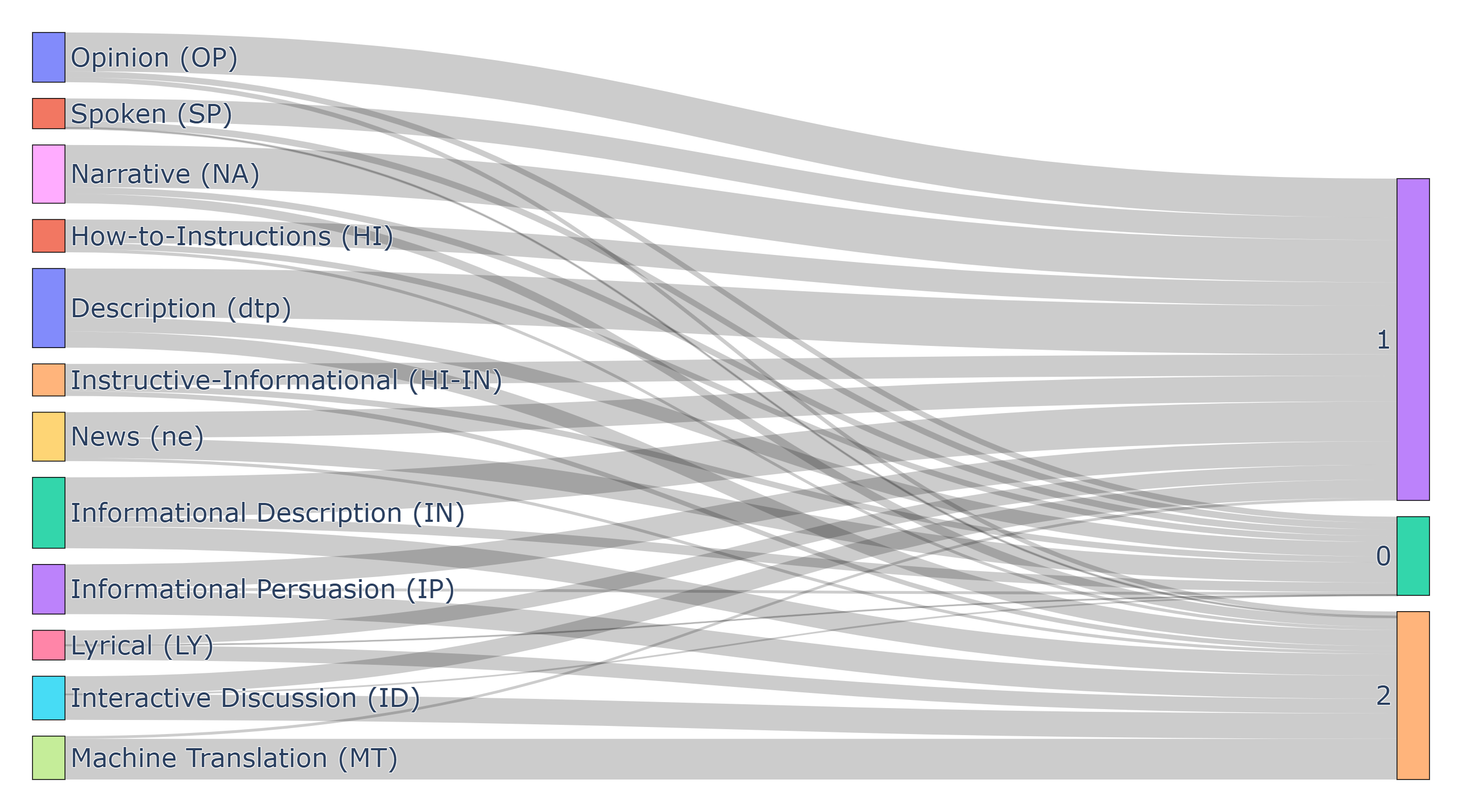}
    \caption{Visualised cross-classification for our register scheme and the NVIDIA NemoCurator classifier. NemoCurator label 0 corresponds to the highest quality and label 2 to the lowest quality.}
    \label{fig:sankey-nvidia}
\end{figure}

Although the above analysis shows associations between the studied schemes, the register classes coincide with all classes of the three tools, with the exception of FineWeb-edu's highest quality class 5, which only occurs in conjunction with register classes \HIIN and \IN. We visualise this in Figures \ref{fig:sankey-dclm}, \ref{fig:sankey-edu}, and \ref{fig:sankey-nvidia}. This means that our register scheme can be seen as mutually reinforcing when combined with other established quality labelling schemes. We specifically see from our results that the 3 tools we compared to the register scheme often assign high-quality labels to documents from registers \IN, \DTP, and the case of NemoCurator, \NE, but neglect classes that we found beneficial in this study, such as \HI and \OP. This highlights the value of using the non-binary but linguistically motivated and versatile scheme that the registers offer for curating data.


\section{Numerical results}
\label{sec:appendix_numerical}

We present the results in numerical format in Table~\ref{tab:appendix-numerical-results}. As stated in Section \ref{sec:evaluation}, we reweigh the accuracy outputted by \texttt{Lighteval}, \texttt{acc\_norm}, as the calculation of \texttt{acc\_norm} is heavily biased against MLLU. MMLU consists of 57 tasks, which are considered separate benchmarks by \texttt{LightEval}. This is why values of \texttt{acc\_norm} are lower than \quot{Accuracy} in the table: average performance of our models on MMLU is just under 0.3.

\begin{table}[!b]
\centering
\small
\begin{tabular}{@{}llllll@{}}
\toprule
Pretraining dataset                                            & Step  & Tokens & Accuracy & \texttt{acc\_norm} & \texttt{stderr}   \\ \midrule
\multicolumn{1}{l|}{\multirow{2}{*}{FineWeb}}                  & 1000  & 2 B    & 0.348450            & 0.269707             & 0.030520 \\
\multicolumn{1}{l|}{}                                          & 50000 & 105 B  & 0.465567            & 0.330767             & 0.031929 \\ \midrule
\multicolumn{1}{l|}{\multirow{2}{*}{\HIa}}                       & 1000  & 2 B    & 0.351006            & 0.264036             & 0.030214 \\
\multicolumn{1}{l|}{}                                          & 50000 & 105 B  & 0.446558            & 0.294467             & 0.030880 \\ \midrule
\multicolumn{1}{l|}{\multirow{2}{*}{\HIINa}}                    & 1000  & 2 B    & 0.352893            & 0.266001             & 0.030249 \\
\multicolumn{1}{l|}{}                                          & 50000 & 105 B  & 0.464502            & 0.314607             & 0.031506 \\ \midrule
\multicolumn{1}{l|}{\multirow{2}{*}{\Mthree}}             & 1000  & 2 B    & 0.354339            & 0.265349             & 0.030291 \\
\multicolumn{1}{l|}{}                                          & 50000 & 105 B  & 0.465616            & 0.320815             & 0.031642 \\ \midrule
\multicolumn{1}{l|}{\multirow{2}{*}{\Mfour}}          & 1000  & 2 B    & 0.352104            & 0.267529             & 0.030357 \\
\multicolumn{1}{l|}{}                                          & 50000 & 105 B  & 0.475093            & 0.331857             & 0.032058 \\ \midrule
\multicolumn{1}{l|}{\multirow{2}{*}{\Mfive}}       & 1000  & 2 B    & 0.354445            & 0.264463             & 0.030214 \\
\multicolumn{1}{l|}{}                                          & 50000 & 105 B  & 0.472279            & 0.319987             & 0.031603 \\ \midrule
\multicolumn{1}{l|}{\multirow{2}{*}{\Msix}}    & 1000  & 2 B    & 0.348062            & 0.263200             & 0.030155 \\
\multicolumn{1}{l|}{}                                          & 50000 & 105 B  & 0.463591            & 0.324652             & 0.031859 \\ \midrule
\multicolumn{1}{l|}{\multirow{2}{*}{\Mseven}} & 1000  & 2 B    & 0.353098            & 0.268480             & 0.030369 \\
\multicolumn{1}{l|}{}                                          & 50000 & 105 B  & 0.464738            & 0.325085             & 0.031820 \\ \midrule
\multicolumn{1}{l|}{\multirow{2}{*}{HPLT v2}}                  & 1000  & 2 B    & 0.341536            & 0.265107             & 0.030274 \\
\multicolumn{1}{l|}{}                                          & 50000 & 105 B  & 0.457484            & 0.328140             & 0.032064 \\ \midrule
\multicolumn{1}{l|}{\multirow{2}{*}{\IDa}}                       & 1000  & 2 B    & 0.336598            & 0.264045             & 0.030252 \\
\multicolumn{1}{l|}{}                                          & 50000 & 105 B  & 0.430686            & 0.298429             & 0.031214 \\ \midrule
\multicolumn{1}{l|}{\multirow{2}{*}{\INa}}                       & 1000  & 2 B    & 0.344821            & 0.270357             & 0.030517 \\
\multicolumn{1}{l|}{}                                          & 50000 & 105 B  & 0.437347            & 0.333771             & 0.032217 \\ \midrule
\multicolumn{1}{l|}{\multirow{2}{*}{\IPa}}                       & 1000  & 2 B    & 0.336226            & 0.261104             & 0.030146 \\
\multicolumn{1}{l|}{}                                          & 50000 & 105 B  & 0.393026            & 0.276895             & 0.030480 \\ \midrule
\multicolumn{1}{l|}{\multirow{2}{*}{\LYa}}                       & 1000  & 2 B    & 0.322654            & 0.261681             & 0.030248 \\
\multicolumn{1}{l|}{}                                          & 50000 & 105 B  & 0.357626            & 0.271479             & 0.030604 \\ \midrule
\multicolumn{1}{l|}{\multirow{2}{*}{\MTa}}                       & 1000  & 2 B    & 0.331829            & 0.262588             & 0.030269 \\
\multicolumn{1}{l|}{}                                          & 50000 & 105 B  & 0.374349            & 0.281125             & 0.030821 \\ \midrule
\multicolumn{1}{l|}{\multirow{2}{*}{\NAa}}                       & 1000  & 2 B    & 0.336791            & 0.263844             & 0.030236 \\
\multicolumn{1}{l|}{}                                          & 50000 & 105 B  & 0.441469            & 0.297377             & 0.031149 \\ \midrule
\multicolumn{1}{l|}{\multirow{2}{*}{\OPa}}                       & 1000  & 2 B    & 0.343446            & 0.262086             & 0.030109 \\
\multicolumn{1}{l|}{}                                          & 50000 & 105 B  & 0.446629            & 0.318284             & 0.031720 \\ \midrule
\multicolumn{1}{l|}{\multirow{2}{*}{\SPa}}                       & 1000  & 2 B    & 0.340778            & 0.266882             & 0.030336 \\
\multicolumn{1}{l|}{}                                          & 50000 & 105 B  & 0.422314            & 0.303859             & 0.031381 \\ \midrule
\multicolumn{1}{l|}{\multirow{2}{*}{\DTPa}}                      & 1000  & 2 B    & 0.347940            & 0.268266             & 0.030463 \\
\multicolumn{1}{l|}{}                                          & 50000 & 105 B  & 0.452347            & 0.329816             & 0.032027 \\ \midrule
\multicolumn{1}{l|}{\multirow{2}{*}{\NEa}}                       & 1000  & 2 B    & 0.332720            & 0.257563             & 0.029931 \\
\multicolumn{1}{l|}{}                                          & 50000 & 105 B  & 0.417933            & 0.304612             & 0.031457 \\ \bottomrule
\end{tabular}
\caption{Accuracies for all models, first and last checkpoint. \quot{Accuracy} denotes accuracy weighted by benchmark, which we present in our figures. Reweighing was done to mitigate the dominance of MMLU in the results, see Section \ref{sec:evaluation}. \texttt{acc\_norm} and \texttt{stderr} stand for average accuracy and standard error as given by \texttt{LightEval}.}
\label{tab:appendix-numerical-results}
\end{table}

\section{Performance by benchmark and register}
\label{sec:app-detailed}

\begin{figure}[!ht]
    \small
    \centering
    \includegraphics[width=\linewidth]{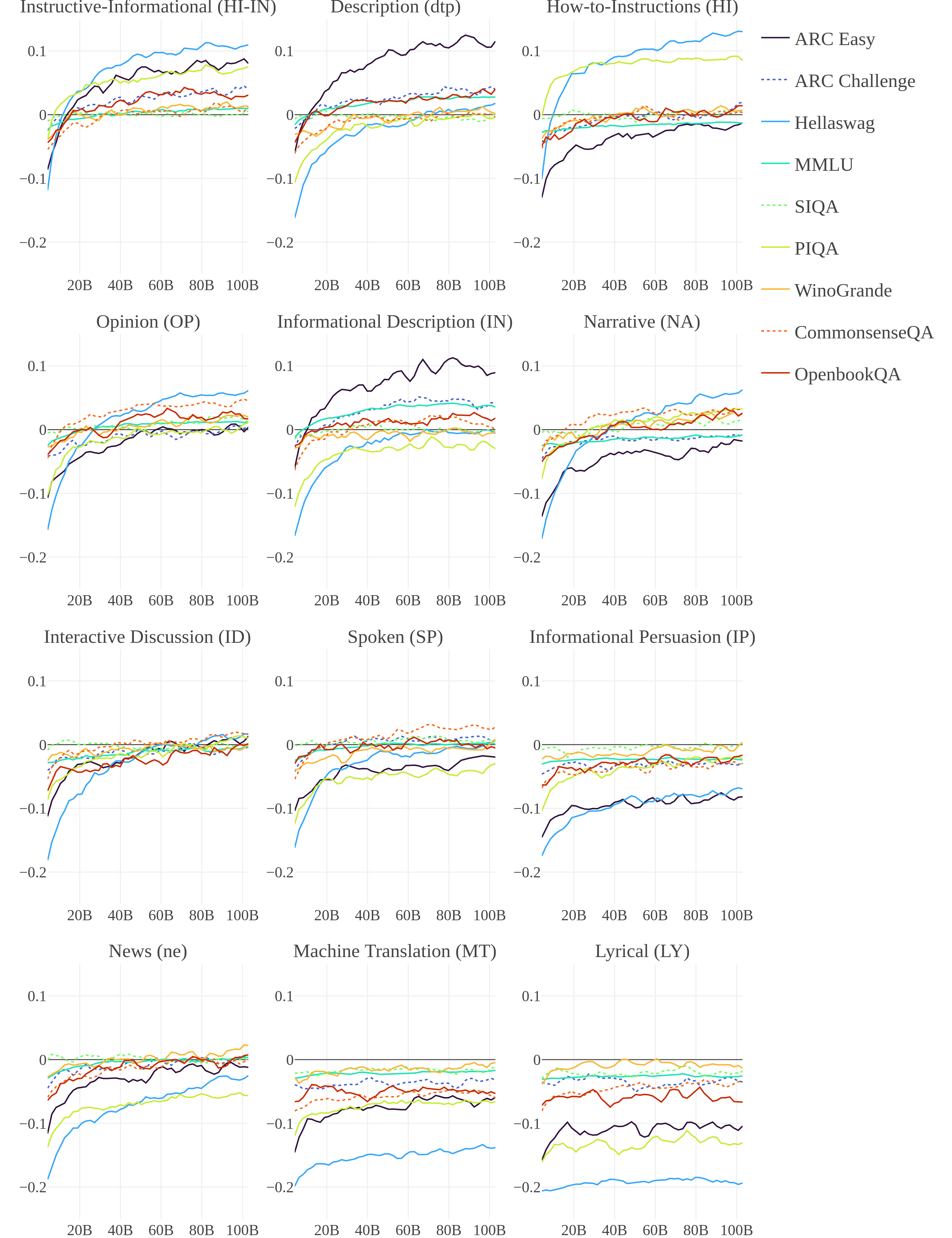}
    \caption{Models' performance by register. Results are scaled by the average last checkpoint performance to highlight the models' differences. Dotted lines to increase readability.}
    \label{fig:appendix-by-register}
\end{figure}

\begin{figure}[!ht]
    \small
    \includegraphics[width=\linewidth]{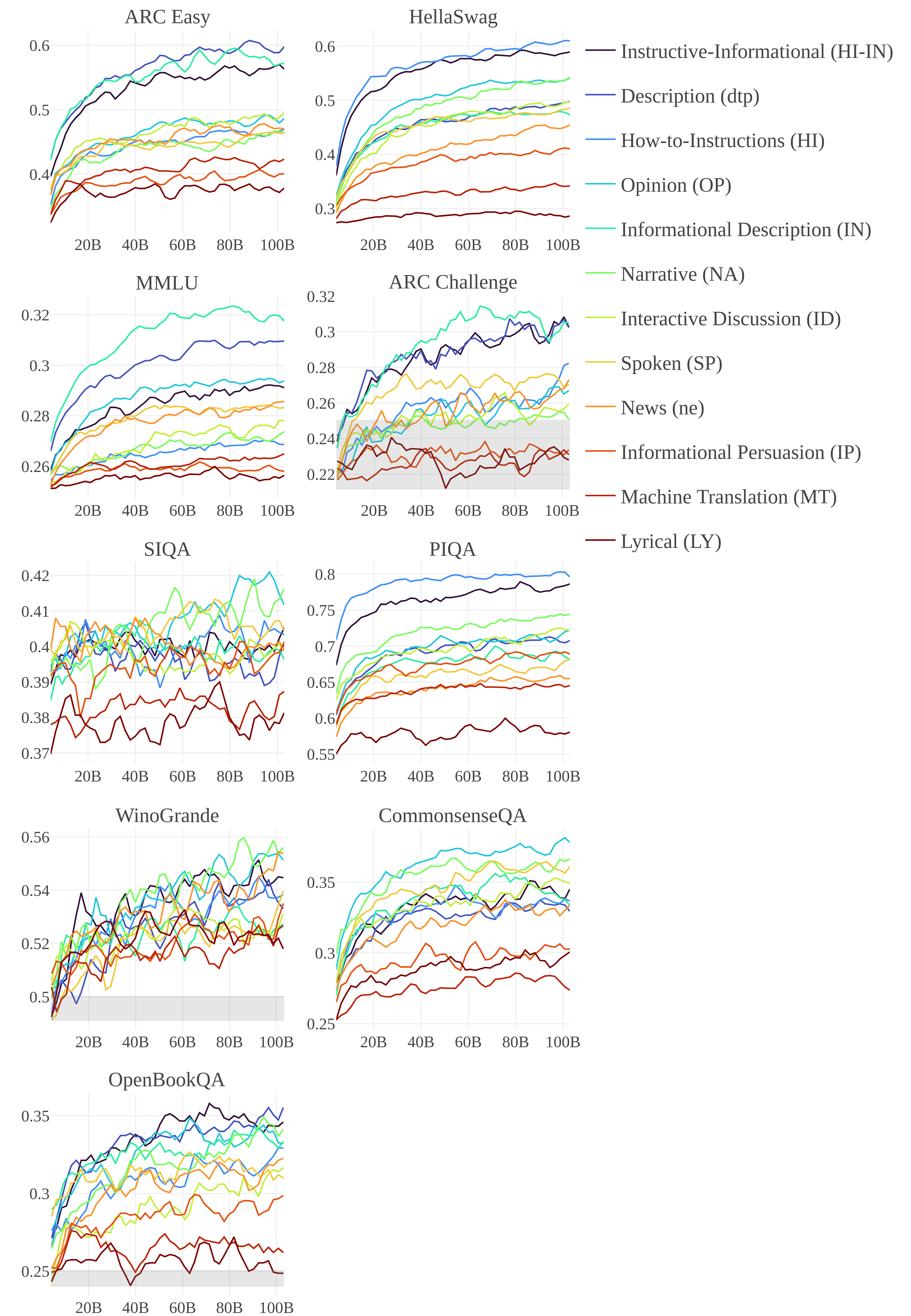}
    \caption{Accuracies of register models on individual benchmarks. Random-guess thresholds shown in grey when at least one model falls under it. Rolling average over 6 billion tokens applied for ease of reading overlapping lines.}
    \label{fig:appendix-by-benchmark-individual}
\end{figure}

\begin{figure}[!ht]
    \small
    \includegraphics[width=\linewidth]{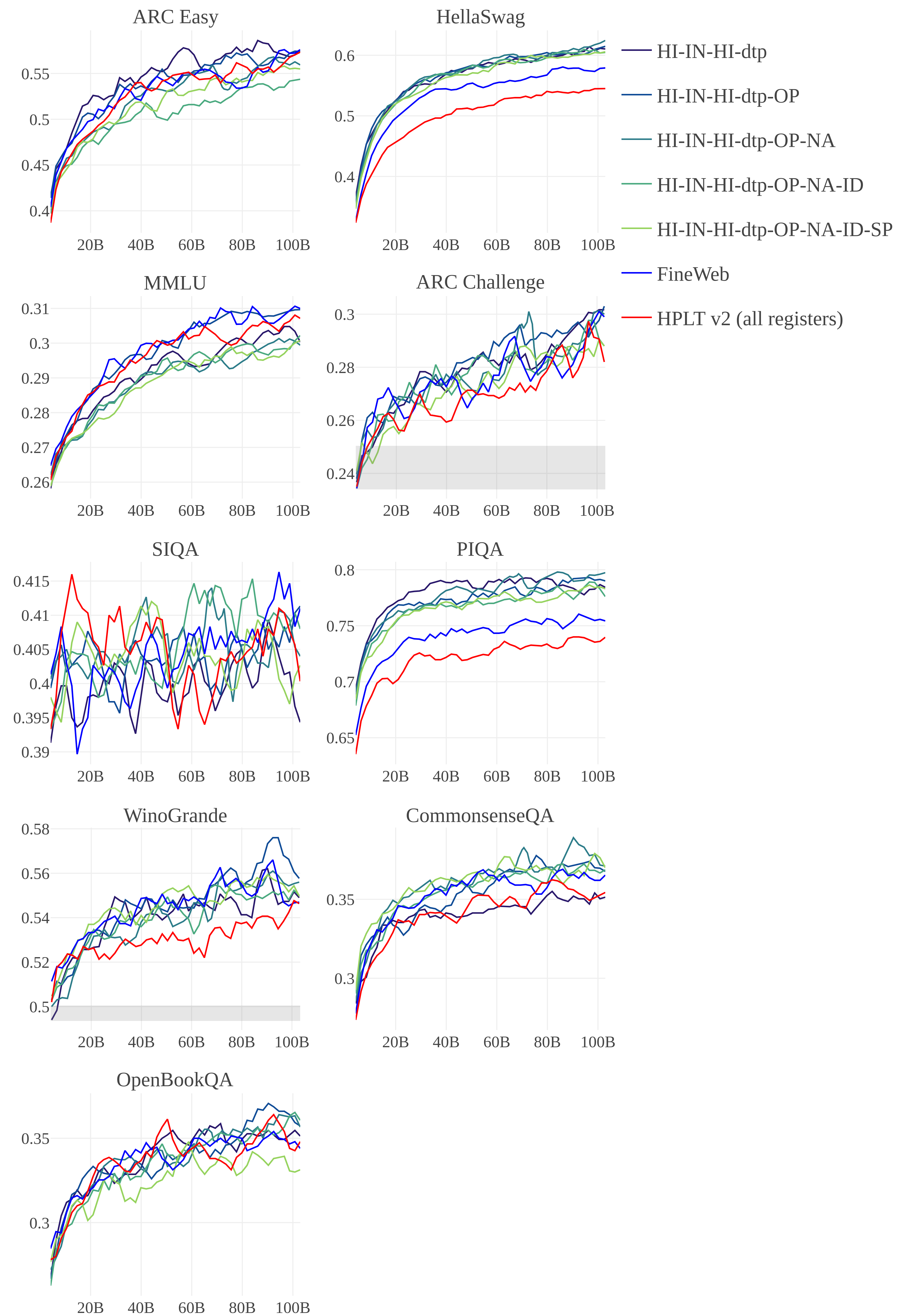}
    \caption{Accuracies of combination models on individual benchmarks. Random-guess thresholds shown in grey when at least one model falls under it. Rolling average over 6 billion tokens applied for ease of reading overlapping lines.}
    \label{fig:appendix-by-benchmark-combination}
\end{figure}

Figures~\ref{fig:appendix-by-register}, \ref{fig:appendix-by-benchmark-individual}, and \ref{fig:appendix-by-benchmark-combination} present the results over all checkpoints, with Figure~\ref{fig:appendix-by-register} showing results by register, others by benchmark. 
To augment readability, we again applied a rolling average. In Figure~\ref{fig:appendix-by-register}, the values are normalised with respect to the average benchmark score on the final checkpoint; for example, the \DTPa model outperforms the average of all registers by 0.1 points on the final checkpoint on the \textit{ARC Easy} benchmark.

\end{document}